\documentclass{article} %
\usepackage[table]{xcolor}
\usepackage{iclr2026_conference,times}
\usepackage[margin=1in]{geometry}

\usepackage{amsmath,amsfonts,bm}

\def\eqref#1{equation~\ref{#1}}

\def\1{\bm{1}}

\DeclareMathAlphabet{\mathsfit}{\encodingdefault}{\sfdefault}{m}{sl}
\SetMathAlphabet{\mathsfit}{bold}{\encodingdefault}{\sfdefault}{bx}{n}

\usepackage{tikz}
\usepackage{graphicx}
\usetikzlibrary{calc}
\usetikzlibrary{arrows.meta} 
\usepackage{enumitem}
\usetikzlibrary{positioning,matrix}

\usetikzlibrary{positioning}
\usepackage{amssymb}
\usepackage{threeparttable}
\usepackage{pdfpages}

\usepackage{pgfmath}
\usepackage{booktabs}   %
\usepackage{siunitx}    %
\usepackage{multirow}   %
\usepackage{graphicx}   %

\newcommand{\cmark}{\checkmark}
\newcommand{\xmark}{\text{\sffamily X}}
\newcommand{\pmark}{\(\sim\)} %

\usepackage{graphicx,subcaption,tikz,amsmath}
\usetikzlibrary{calc,positioning}

\definecolor{iceBlue}{HTML}{E5ECF8}
\definecolor{chilledyellow}{HTML}{FFD59C}
\definecolor{mintMist}{HTML}{edd1e7}
\definecolor{easygreen}{HTML}{c3dde4}
\definecolor{easypurple}{HTML}{B6C7FF}
\definecolor{darkpurple}{HTML}{ace3b8}
\definecolor{violetnew}{HTML}{FF2052}
\definecolor{cocieles}{HTML}{872657}

\usepackage{pgfplots}
\pgfplotsset{compat=1.17}

\usepackage{pgfmath, pgffor}

\usepackage{tikz} %

\definecolor{perfLow}{HTML}{DDDDDD}   %
\definecolor{perfHigh}{HTML}{010082}  %

\def\rowmin{0}
\def\rowmax{1}

\newcommand{\scanrow}[1]{%
    \gdef\rowmin{100}%
    \gdef\rowmax{-100}%
    \foreach \x in {#1} {%
        \pgfmathparse{\x < \rowmin ? \x : \rowmin}\global\let\rowmin\pgfmathresult%
        \pgfmathparse{\x > \rowmax ? \x : \rowmax}\global\let\rowmax\pgfmathresult%
    }%
}

\newcommand{\perfcellrev}[1]{%
    \pgfmathparse{(\rowmax - \rowmin) == 0 ? 0 : 1}%
    \ifnum\pgfmathresult=0
        #1%
    \else
        \pgfmathsetmacro{\norm}{(#1 - \rowmin) / (\rowmax - \rowmin)}%
        \pgfmathparse{\norm < 0 ? 0 : (\norm > 1 ? 1 : \norm)}%
        \let\norm\pgfmathresult
        \pgfmathsetmacro{\normrev}{1 - \norm}%
        \pgfmathtruncatemacro{\perc}{20 + 80 * \normrev}%
        \textcolor{perfHigh!\perc!perfLow}{#1}%
    \fi
}
\newcommand{\perfcell}[1]{%
    \pgfmathparse{(\rowmax - \rowmin) == 0 ? 0 : 1}%
    \ifnum\pgfmathresult=0
        #1%
    \else
        \pgfmathsetmacro{\norm}{(#1 - \rowmin) / (\rowmax - \rowmin)}%
        \pgfmathparse{\norm < 0 ? 0 : (\norm > 1 ? 1 : \norm)}%
        \let\norm\pgfmathresult
        \pgfmathtruncatemacro{\perc}{20 + 80 * \norm}%
        \textcolor{perfHigh!\perc!perfLow}{#1}%
    \fi
}

\def\perfcellaux#1.#2\relax{%
  \ifnum#1=1
    \def\perc{100}%
  \else
    \edef\perc{#2}%
  \fi
  \cellcolor{perfHigh!\perc!perfLow}{#1.#2}%
}

\tikzset{
  card/.style={rounded corners=8pt, line width=0.5pt, inner sep=6pt},
  cardlabel/.style={
    anchor=north west,
    font=\bfseries\small,
    text=orange!70!black,
    xshift=4pt,
    yshift=12pt
  }
}

\newcommand{\sonicpipeline}{%
  $x \xrightarrow{\ \mathcal{F}\ }\widehat{x}
  \xrightarrow{\ \times\,T(\boldsymbol{\omega})\ }
  \widehat{y}
  \xrightarrow{\ \mathcal{F}^{-1}\ } y$}

\usepackage{hyperref}
\usepackage{url}
\usepackage{subcaption}
\usepackage{graphicx}
\usepackage{algorithm}      %
\usepackage{algpseudocode}  %
\usepackage{float}
\usepackage{tikz}
\usetikzlibrary{positioning,fit,patterns,shadings,calc}
\usepackage{pgfplots}
\usepackage{amsmath}

\usepackage{booktabs}
\usepackage{multirow}

\newcommand{\iconIso}{%
  \tikz[baseline=-0.6ex,scale=0.35]{%
    \fill[darkpurple] (0,0) circle (0.12);
  }%
}

\newcommand{\iconAxis}{%
  \tikz[baseline=-0.6ex,scale=0.35]{%
    \draw[easypurple, very thick] (-0.18,0) -- (0.18,0);
    \draw[easypurple, very thick] (0,-0.18) -- (0,0.18);
  }%
}

\newcommand{\iconOrient}{%
  \tikz[baseline=-0.6ex,scale=0.35]{%
    \fill[chilledyellow!90!black] (0,0) circle (0.04);
    \foreach \a in {0,45,...,315}{%
      \draw[chilledyellow!80!black, very thick] (0,0) -- (\a:0.24);
    }
  }%
}

\newcommand{\LRFOTaxonomy}{%
\hspace{-5mm}
\scalebox{0.75}{%
\begin{tikzpicture}[
    font=\small\sffamily,
    box/.style={rounded corners=8pt, draw=black!20, thick, align=center, inner sep=6pt},
    subbox/.style={rounded corners=6pt, draw=black!20, semithick, align=center, inner sep=5pt},
    title/.style={font=\bfseries}
]

\node[box, fill=iceBlue!90, minimum width=18cm, minimum height=2.6cm] (outer) {};
\node[title] at (outer.north) [yshift=-11pt]
    {Large Receptive-Field Operators};

\node[box, fill=easypurple!75, minimum width=266pt, minimum height=2.6cm,
      anchor=west] (rinv) at ([xshift=0.2cm,yshift=-21pt]outer.west) {};
\node[title] at (rinv.north) [yshift=-9pt]
    {Resolution-Invariant};

\node[box, fill=easygreen!75, minimum width=240pt, minimum height=2.6cm,
      anchor=east] (rdep) at ([xshift=0.1cm,yshift=-21pt]outer.east) {};
\node[title] at (rdep.north) [yshift=-9pt]
    {Resolution-Dependent};

\node[box, fill=mintMist!60, minimum width=13.3cm, minimum height=2.6cm,
      anchor=west] (spec) at ([xshift=0.3cm,yshift=-40pt]rinv.west |- outer.center) {};
\node[title] at (spec.north) [yshift=-9pt]
    {Spectral-Domain};

\coordinate (baselineY) at ([yshift=0.30cm]spec.south west);
\coordinate (specSWbase) at ([xshift=0.25cm]baselineY);
\coordinate (specSEbase) at (spec.south east |- baselineY);
\coordinate (rdepSEbase) at (rdep.south east |- baselineY);

\node[subbox, fill=chilledyellow!60,
      text width=111pt,
      font=\footnotesize,
      minimum width=75pt,
      minimum height=87pt,
      anchor=south east]
      (spatial) at ([xshift=3pt,yshift=-21pt]rdepSEbase) {%
    \begin{minipage}{\linewidth}
      \vspace{2mm}
      \begin{itemize}[leftmargin=22pt, nosep]
        \item Large-kernel CNNs
        \item Non-local networks
        \item A$^2$-Nets
        \item Vision transformers
      \end{itemize}
    \end{minipage}
};
\node[title] at (spatial.north) [yshift=-9pt]
    {Spatial-Domain};

\node[subbox, fill=iceBlue!40,
      text width=5cm,
      minimum height=69pt,
      minimum width=258pt,
      anchor=south west]
      (contbox) at ([xshift=0.05cm,yshift=-21pt]specSWbase) {%
};
\node[title] at (contbox.north) [yshift=-9pt]
    {Continuous};

\coordinate (contSWbase) at ([xshift=8pt,yshift=-20pt]contbox.south west);
\coordinate (contSEbase) at ([xshift=1pt,yshift=-20pt]contbox.south east);

\node[subbox, fill=darkpurple!40,
      text width=101pt,
      minimum width=64pt,
      minimum height=69pt,
      anchor=south east]
      (discrete) at ([xshift=2pt,yshift=-21pt]specSEbase) {%
      \begin{minipage}{\linewidth}
      \vspace{2mm}
      \begin{itemize}[leftmargin=22pt, nosep]
        \item FNO
        \item Global Filter
        \item Spectral CNNs
      \end{itemize}
      \end{minipage}
};
\node[title] at (discrete.north) [yshift=-9pt]
    {Discrete Grid};

\node[subbox,
      fill=chilledyellow!20,
      text width=5.32cm,
      minimum height=70pt,
      anchor=south west]
      (structured) at (contSWbase) {%
    \begin{minipage}{\linewidth}
      \centering
      \scriptsize
      \setlength{\tabcolsep}{2pt} %
      \begin{tabular}{@{}lccc@{}}
         & \textbf{SNO} & \textbf{S4ND} & \textbf{\color{blue}SONIC} \\[4pt]
          \small \emph{Orientation-aware} & \xmark & \pmark & \color{blue}\cmark \\[6pt]
          \small \emph{Low-rank}          & \xmark & \xmark & \color{blue}\cmark \\
      \end{tabular}
    \end{minipage}%
};
\node[title] at (structured.north) [yshift=-9pt]
    {Structured};

\coordinate (arrowStart) at ([xshift=0.55cm]structured.south east);
\coordinate (arrowEnd)   at ([xshift=0.55cm]structured.north east);

\draw[->, very thick, black!40] (arrowStart) -- (arrowEnd);
\node at ($(arrowStart)!0.15!(arrowEnd)$) {\iconIso};
\node at ($(arrowStart)!0.50!(arrowEnd)$) {\iconAxis};
\node at ($(arrowStart)!0.85!(arrowEnd)$) {\iconOrient};

\node[subbox, fill=mintMist!20,
      text width=2.7cm,
      minimum width=64pt,
      minimum height=70pt,
      anchor=south east]
      (fully) at (contSEbase) {%
      \begin{minipage}{\linewidth}
      \vspace{2mm}
      \begin{itemize}[leftmargin=22pt, nosep]
        \item NIFF
        \item Continuous FNO
      \end{itemize}
      \end{minipage}
};
\node[title] at (fully.north) [yshift=-9pt]
    {Unstructured};

\end{tikzpicture}%
}%
} %
\newcommand{\AllFigs}{%
  \centering
  \setlength{\fboxsep}{0pt}
  \setlength{\fboxrule}{0pt}

 \begin{subfigure}[t]{\linewidth}
  \centering
  \begin{tikzpicture}
    \node[card,fill=easygreen] (AllCard) {%
      \begin{minipage}{\linewidth}\centering

        \vspace{10pt}
        \begin{minipage}[t]{0.47\linewidth}
          \centering
          \begin{tikzpicture}[remember picture] %
            \node[
              rectangle,
              rounded corners=3pt,
              draw,
              thick,
              fill=mintMist,
              inner sep=4pt
            ] (SFcard) {%
              \begin{minipage}{0.96\linewidth}\centering
                \vspace{6pt}%
                \begin{subfigure}{0.18\linewidth}
                  \includegraphics[width=\linewidth]{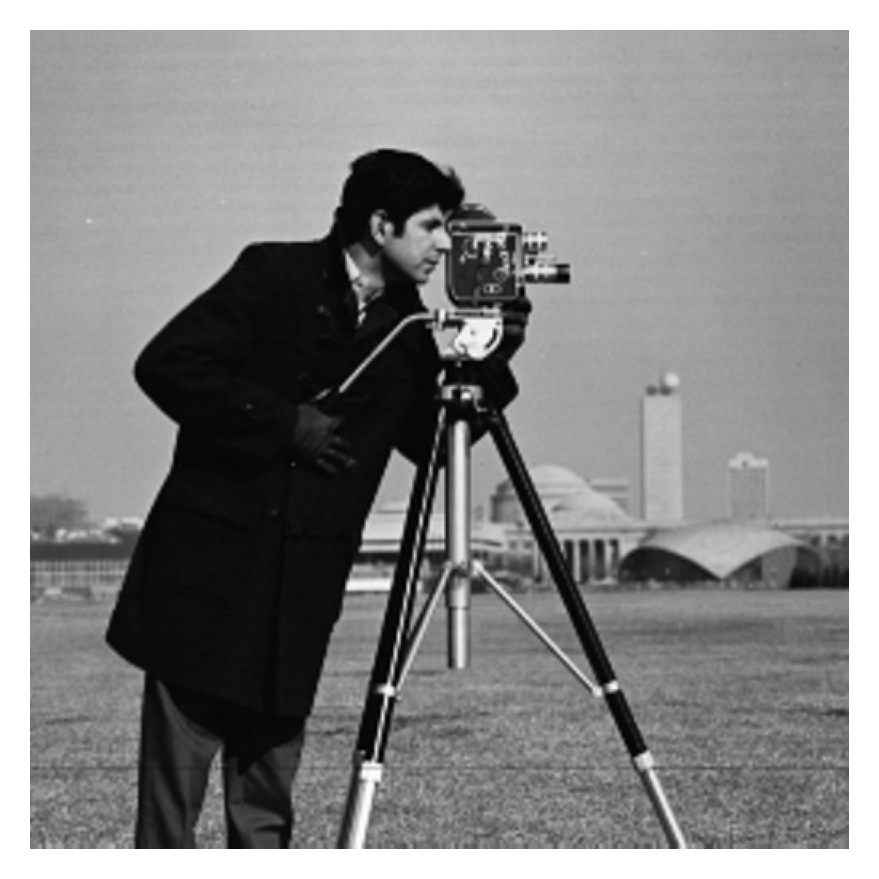}
                  \caption*{\large $x^{(\ell)}$}
                \end{subfigure}\hfill
                \begin{subfigure}{0.18\linewidth}
                  \includegraphics[width=\linewidth]{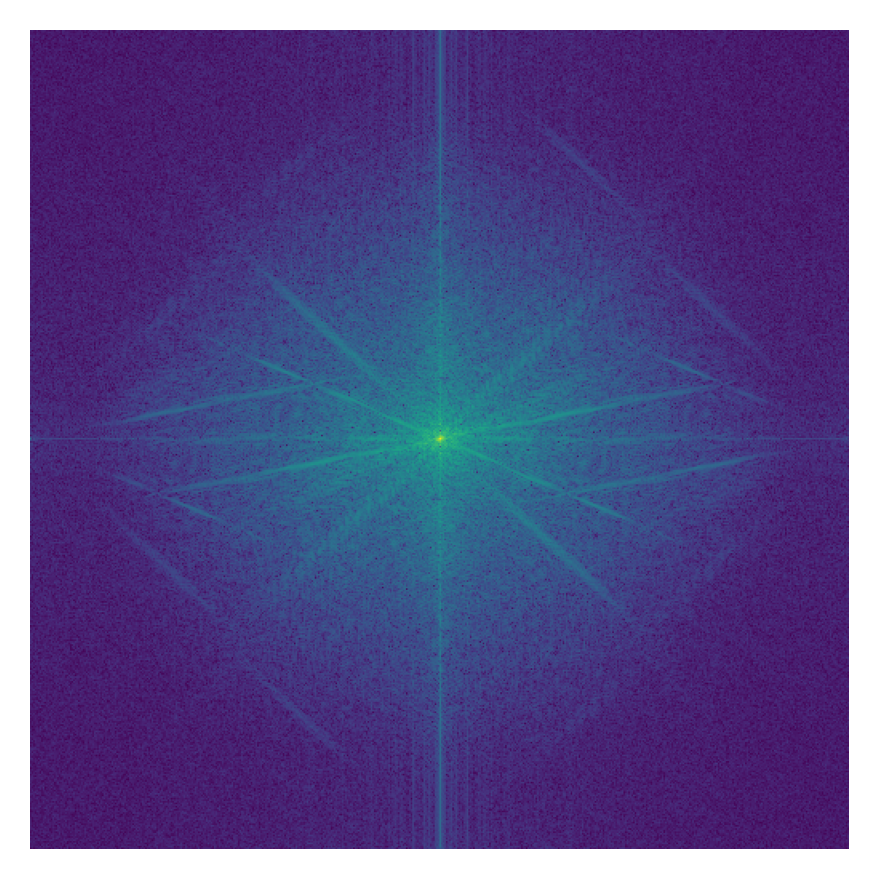}
                  \caption*{\large $\widehat{x}$}
                \end{subfigure}\hfill
                \begin{subfigure}{0.18\linewidth}
                  \begin{tikzpicture}
                    \node[
                      rounded corners=4pt,
                      inner sep=1pt,yshift=-8pt
                    ] (Hmain)
                    {\includegraphics[width=\linewidth]{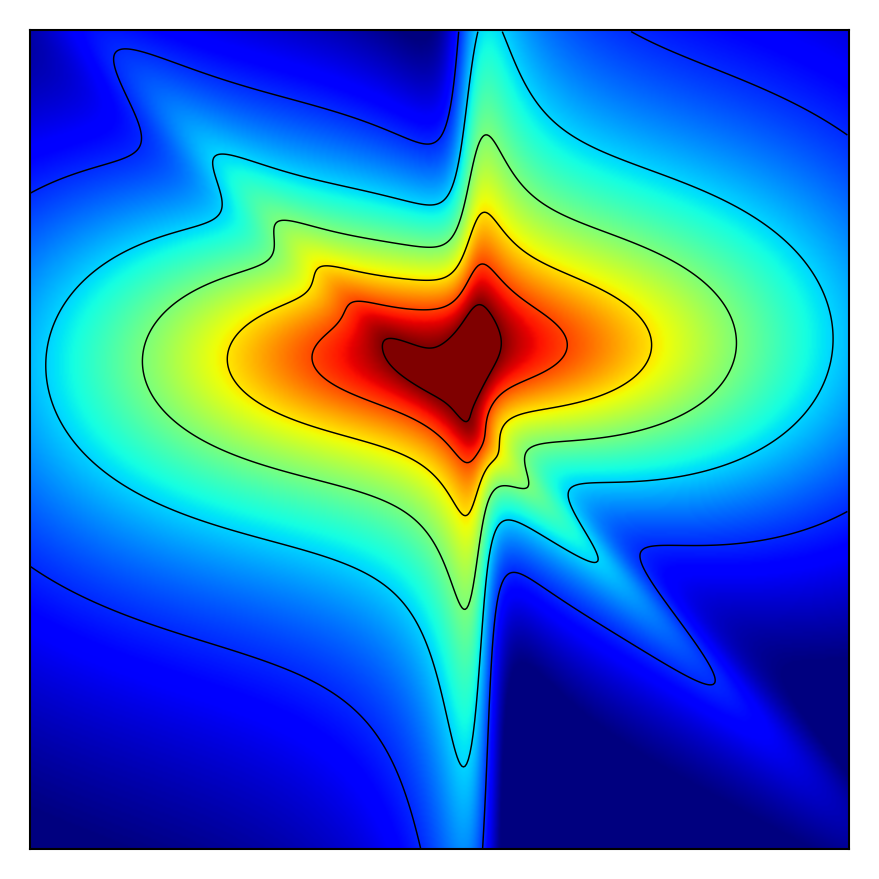}};
                  \end{tikzpicture}
                  \caption*{\large
                    \tikz[baseline=(H.base)]
                      \node[draw, circle, thick, inner sep=2pt, minimum size=3mm, fill=darkpurple]
                        (H) {\small $\widehat{H}$};
                  }
                \end{subfigure}\hfill
                \begin{subfigure}{0.18\linewidth}
                  \begin{tikzpicture}                
                    \node[
                      rounded corners=4pt,
                      inner sep=1pt,
                      fill=white
                    ] (Pmain)
                    {\includegraphics[width=\linewidth]{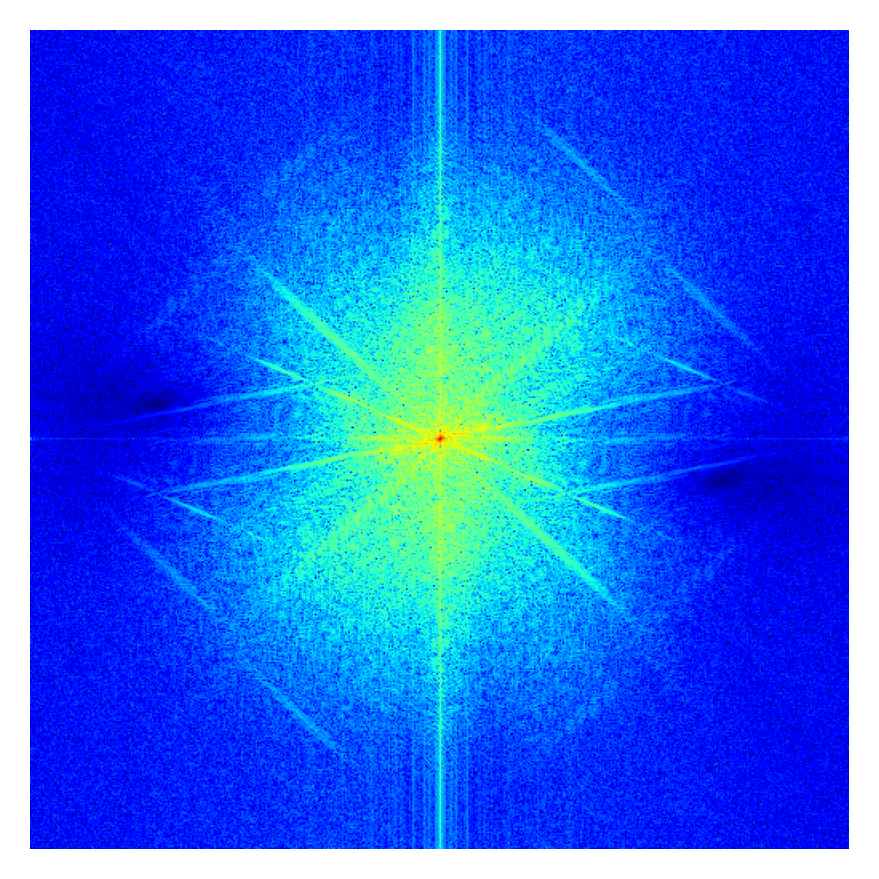}};
                  \end{tikzpicture}
                  \caption*{\large $\widehat{H} \cdot \widehat{x}$}
                \end{subfigure}\hfill
                \begin{subfigure}{0.18\linewidth}
                  \begin{tikzpicture}
                    \node[
                      rounded corners=4pt,
                      inner sep=1pt,
                      fill=white
                    ] (Ymain)
                    {\includegraphics[width=\linewidth]{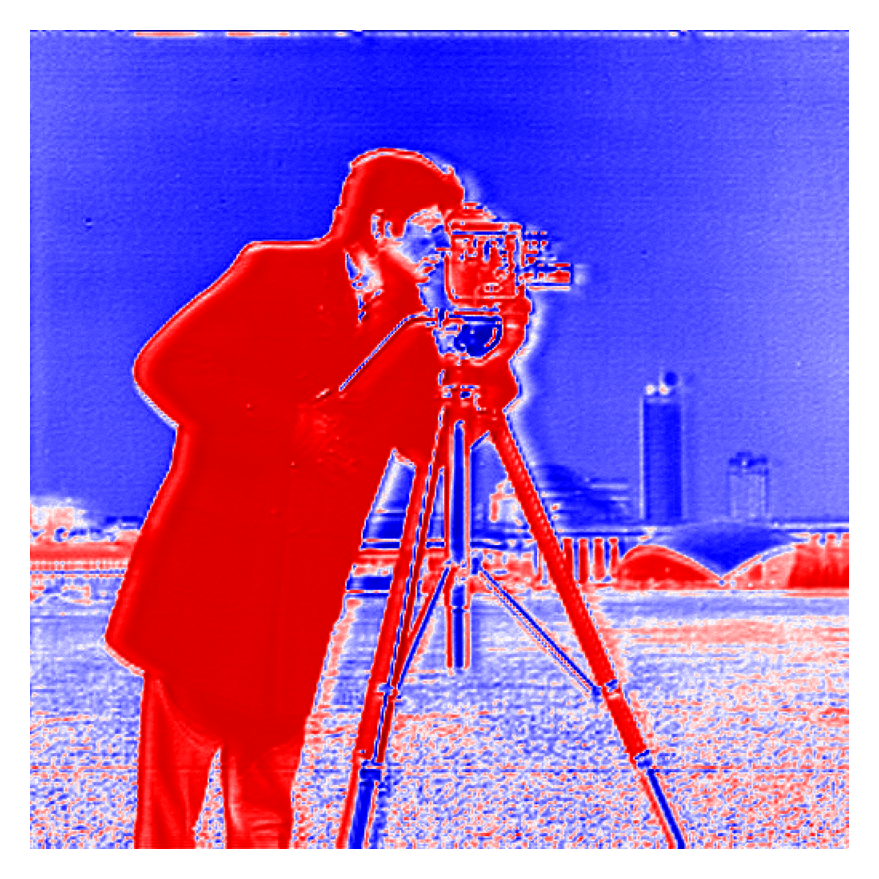}};
                  \end{tikzpicture}
                  \caption*{\large $y^{(\ell)}$}
                \end{subfigure}

                \vspace{1pt}

                {\scriptsize \sonicpipeline}
              \end{minipage}
            };
          \end{tikzpicture}
        \end{minipage}%
        \hfill        %
        \begin{minipage}[t]{0.47\linewidth}
          \centering
            \begin{tikzpicture}[ %
              scale=0.85, 
              every node/.style={transform shape},
              block/.style={
                rectangle,draw,rounded corners=2pt,
                inner sep=1pt,minimum width=0.9cm,minimum height=8mm
              },
              op/.style={circle,draw,inner sep=1pt,minimum size=2mm},
              flow/.style={->,>=Stealth,semithick},
              tinylabel/.style={font=\Large},
              remember picture %
            ]
              \node[coordinate] (xin) {};
              \node[coordinate, below=8mm of xin] (xinbelow) {};
              \node[coordinate, below right=4.5mm and 4mm of xin] (xinright) {};
              
              \node[tinylabel, below=0mm of xin] (xinlabel) {\Large $x^{(\ell)}$};

              \node[block, right=0.5cm of xinlabel, fill=mintMist] (sonic) {SONIC};
              \node[block, right=0.5cm of sonic] (bn) {Norm};
              \node[op, right=0.5cm of bn] (plus) {$+$};
              \node[block, right=0.5cm of plus] (relu) {ReLU};
              \node[tinylabel, right=0.5cm of relu] (xout) {\Large $x^{(\ell+1)}$};

              \draw[flow] (xinright) -- (sonic);
              \draw[flow] (sonic) -- (bn);
              \draw[flow] (bn) -- (plus);

              \node[block,below=0.7cm of bn] (proj) {Proj};
              \draw[flow] (xinbelow) |- (proj.west);
              \draw[flow] (proj) -| (plus.south);

              \draw[flow] (plus) -- (relu);
              \draw[flow] (relu) -- (xout);

              \node[coordinate, above left=15mm and -1mm of relu] (smallanchor) {};
              \node[coordinate, left=-4mm of smallanchor] (smallanchorextra) {};

              \begin{scope}[overlay,yshift=-3mm] %
                \node[font=\large] (xL) at (smallanchor) {$x^{(L)}$};

                \node[block, font=\large, minimum width=0.8cm, minimum height=8mm,
                      right=0.35cm of xL] (linear) {Linear};

                \node[coordinate,right=0.5cm of linear] (yextra) {};
                \node[font=\large, minimum width=1.2cm, minimum height=8mm, right=0.1cm of linear] (yemb) {$\tilde{\boldsymbol{y}}$};

                \draw[flow] (smallanchorextra) -- (linear.west);
                \draw[flow] (linear.east) -- (yextra.west);
              \end{scope}

            \end{tikzpicture}
          \vspace{2pt}
        \end{minipage}

      \end{minipage}
    };

    \node[anchor=north, font=\bfseries\footnotesize]
      at ([yshift=-2pt]AllCard.north) {SONIC Block};
  \end{tikzpicture}

  \begin{tikzpicture}[remember picture,overlay]
    \draw[-,>=Stealth,semithick]
      (sonic.north west) -- (SFcard.north east);
    \draw[-,>=Stealth,semithick]
      (sonic.south west) -- (SFcard.south east);
  \end{tikzpicture}
  \vspace{-5mm}
  \label{fig:sonic_block_pipeline}
\caption{A SONIC block applies a learned frequency response $\widehat{H}(\omega)$ to the input: the feature map is transformed to the 
Fourier domain, modulated by $\widehat{H}$, and returned to the spatial 
domain before normalization and a residual ReLU fusion.}
\end{subfigure} 
 \begin{subfigure}[t]{\linewidth}
  \centering
  \begin{tikzpicture}
    \node[card,fill=easypurple,minimum height=42pt] (CardAll) {%
      \begin{minipage}{\linewidth}\centering
        \vspace{6pt}

        \begin{subfigure}[t]{0.47\linewidth}
  \centering
  \begin{tikzpicture}[x=1.4cm,y=1.0cm]
  \tikzset{
    nodecircle/.style={
      circle,draw,thick,
      inner sep=1pt,
      minimum size=8mm
    },
    chnode/.style={nodecircle,fill=white},
    modenode/.style={nodecircle,fill=chilledyellow},
    outnode/.style={nodecircle,fill=darkpurple},
    flow/.style={->,>=Stealth,semithick},
    tinylabel/.style={font=\scriptsize}
  }

  \node[chnode,fill=iceBlue] (xone)  at (0, 0.7) {$\widehat{x}_1$};
  \node         (xdots) at (0, 0.1) {$\vdots$};
  \node[chnode,fill=iceBlue] (xCin)  at (0,-0.7) {$\widehat{x}_{C}$};

  \node[modenode] (m1) at (1.6,  1.0) {$T_1$};
  \node[modenode] (m2) at (1.6,  0.0) {$T_2$};
  \node[modenode] (m3) at (1.6, -1.0) {$T_3$};

  \node[outnode] (y1)    at (3.2, 0.7) {$\widehat{H}_1$};
  \node           (ydots) at (3.2, 0.1) {$\vdots$};
  \node[outnode] (yCout) at (3.2,-0.7) {$\widehat{H}_K$};

  \foreach \src in {xone,xCin}{
    \foreach \j in {1,2,3}{
      \draw[flow] (\src.east) -- (m\j.west);
    }
  }

  \foreach \j in {1,2,3}{
    \foreach \dst in {y1,yCout}{
      \draw[flow] (m\j.east) -- (\dst.west);
    }
  }

  \coordinate (inMid)   at ($(xone)!0.5!(xCin)$);
  \coordinate (modeMid) at ($(m1)!0.5!(m3)$);
  \coordinate (outMid)  at ($(y1)!0.5!(yCout)$);

  \node[tinylabel,above=0.7cm] at ($(inMid)!0.5!(modeMid)$)
    {$B\in\mathbb{C}^{M \times C}$};
  \node[tinylabel,above=0.7cm] at ($(modeMid)!0.5!(outMid)$)
    {$C\in\mathbb{C}^{K \times M}$};

\end{tikzpicture}

\end{subfigure}
        \begin{subfigure}[t]{0.47\linewidth}
          \centering
          \begin{tikzpicture}
            \node[inner sep=0pt,yshift=8mm] (img1) at (0,0)
              {\includegraphics[width=.3\linewidth]{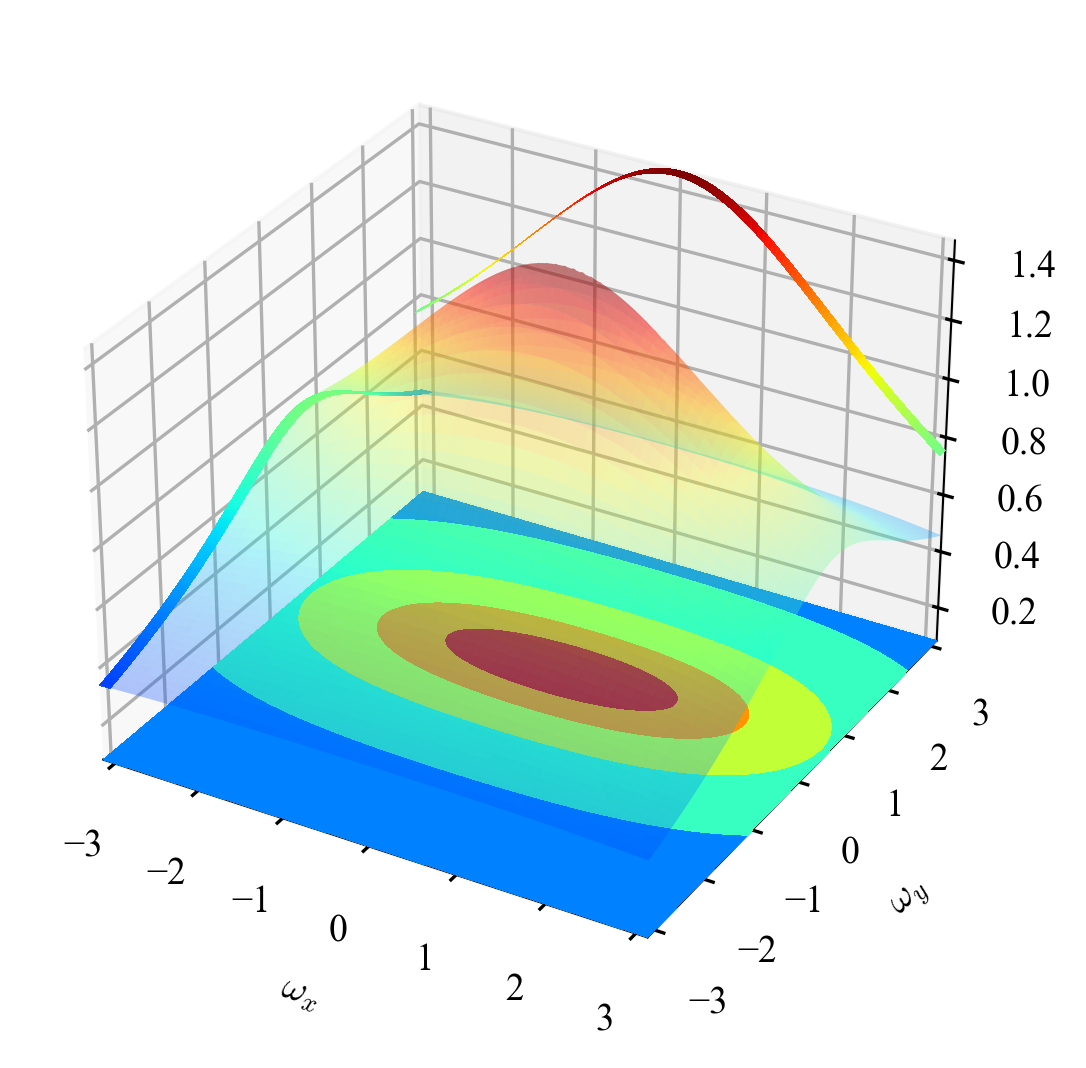}};
            \node[circle,fill=chilledyellow,draw,thick,inner sep=1.2pt,below=2pt of img1,minimum size=7mm,yshift=-2mm]
              {$T_1$};
            \hfill
            \hspace{5mm}
            \node[inner sep=0pt,yshift=8mm] (img2) at (1.9,0)
              {\includegraphics[width=.3\linewidth]{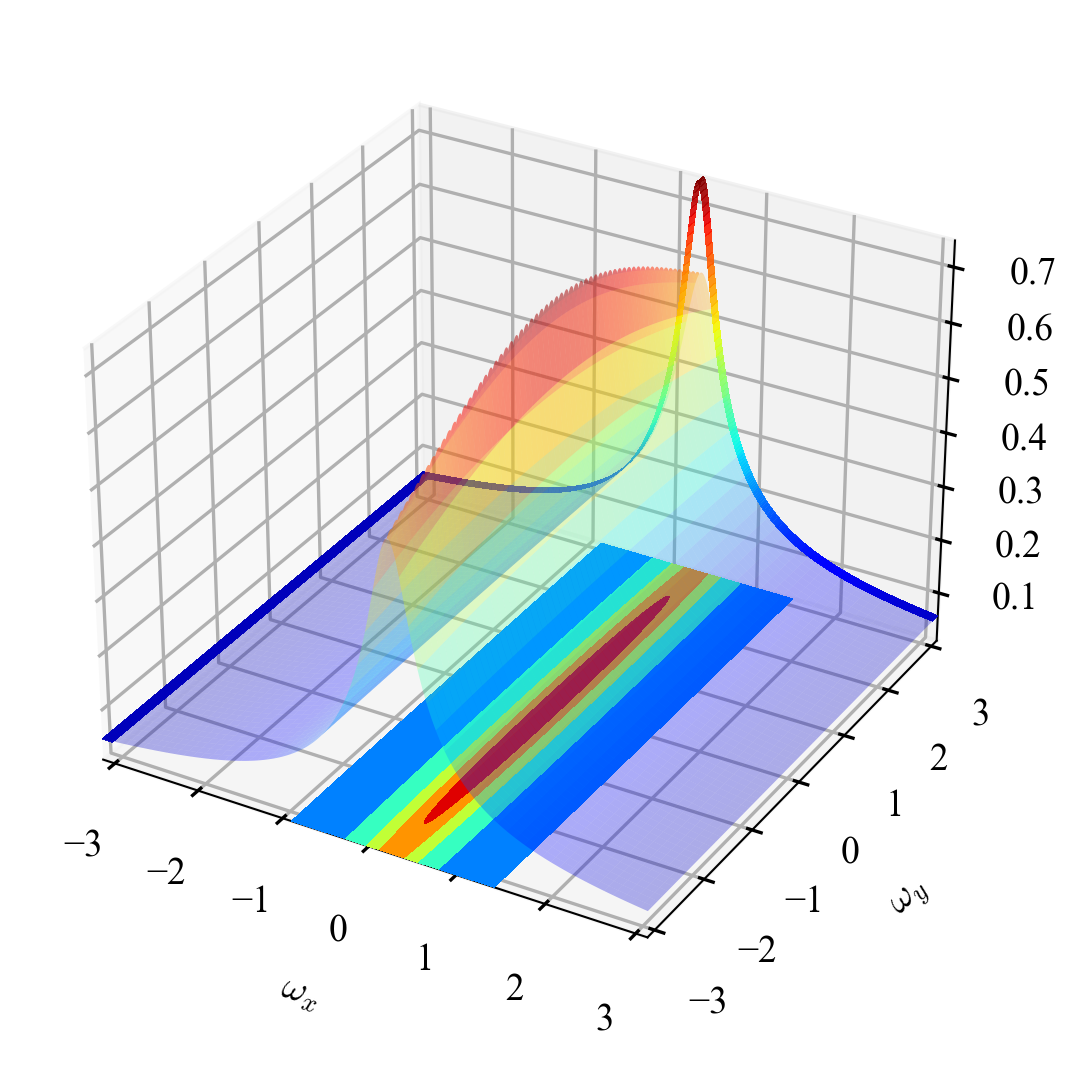}};
            \node[circle,fill=chilledyellow,draw,thick,inner sep=1.2pt,below=2pt of img2,minimum size=7mm,yshift=-2mm]
              {$T_2$};
            \hfill \hspace{5mm}
            \node[inner sep=0pt,yshift=8mm] (img3) at (3.8,0)
              {\includegraphics[width=.3\linewidth]{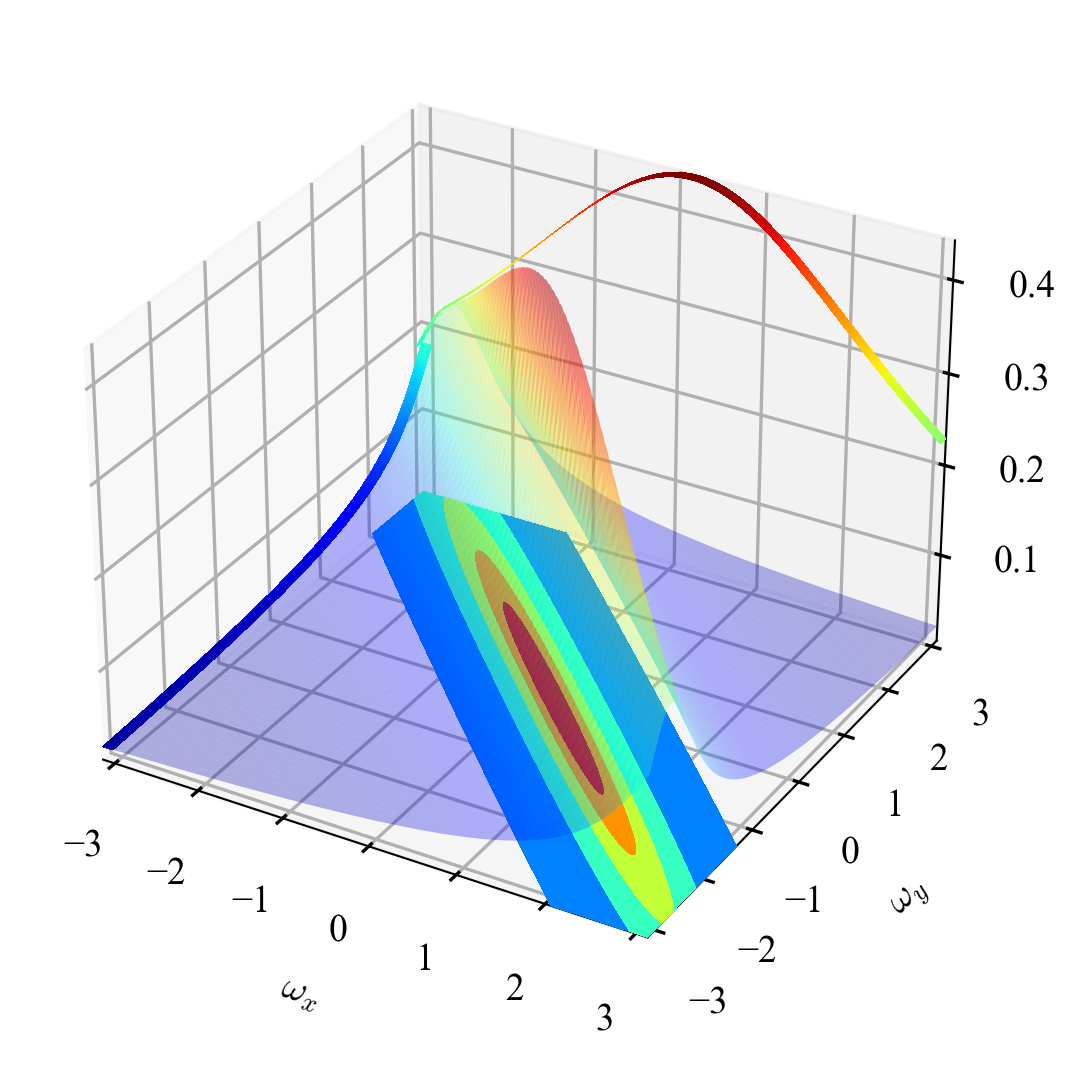}};
            \node[circle,fill=chilledyellow,draw,thick,inner sep=1.2pt,below=2pt of img3,minimum size=7mm,yshift=-2mm]
              {$T_3$};
          \end{tikzpicture}
          \caption*{Mode decomposition}
        \end{subfigure}

      \end{minipage}
    };

    \node[anchor=north, font=\bfseries\footnotesize]
      at ([yshift=-2pt]CardAll.north) {Spectral Mixing};

  \end{tikzpicture}
\caption{The spectral symbol $\widehat{H}(\omega)$ is constructed as a 
superposition of $M$ spectral modes. Each mode $T_m(\omega)$ is a learned 
complex filter over frequency. Input channels are mixed into the modes via 
$B\!\in\!\mathbb{C}^{M\times C}$, and the mode responses are recombined into 
$K$ output channels via $C\!\in\!\mathbb{C}^{K\times M}$, yielding the 
low-rank factorization 
$\widehat{H}_{k,c}(\boldsymbol{\omega})
=\sum_{m=1}^M C_{km}\,T_m(\boldsymbol{\omega})\,B_{mc}$.}
\end{subfigure}
\begin{subfigure}[t]{\linewidth}
  \centering
  \begin{tikzpicture}
    \node[card,minimum height=20pt, fill=chilledyellow] (Dcard) {%
      \begin{minipage}{\linewidth}\centering
        \vspace{12pt}%
        \begin{minipage}{\linewidth}\centering
          \setlength{\tabcolsep}{2pt}
          \begin{tabular}{cccccc}
            \begin{minipage}{0.16\linewidth}
              \centering
              \includegraphics[width=\linewidth]{images/sonic_cam_param_base_3d.png}\\[-1pt]
              {\footnotesize Base kernel}
            \end{minipage}
            &
            \begin{minipage}{0.16\linewidth}
              \centering
              \includegraphics[width=\linewidth]{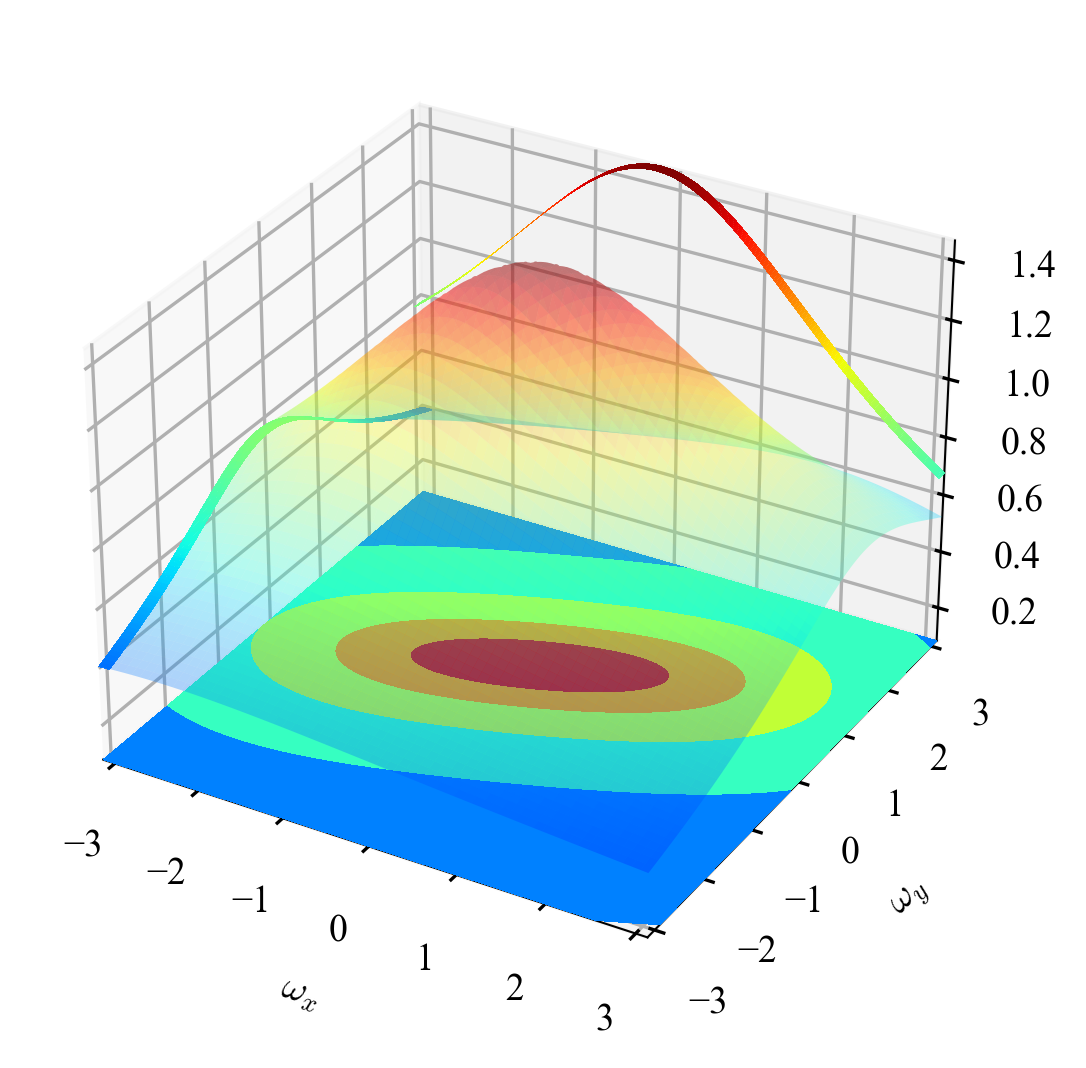}\\[-1pt]
              {\footnotesize Orientation: \textcolor{blue}{$\boldsymbol{v}_1$}}
            \end{minipage}
            &
            \begin{minipage}{0.16\linewidth}
              \centering
              \includegraphics[width=\linewidth]{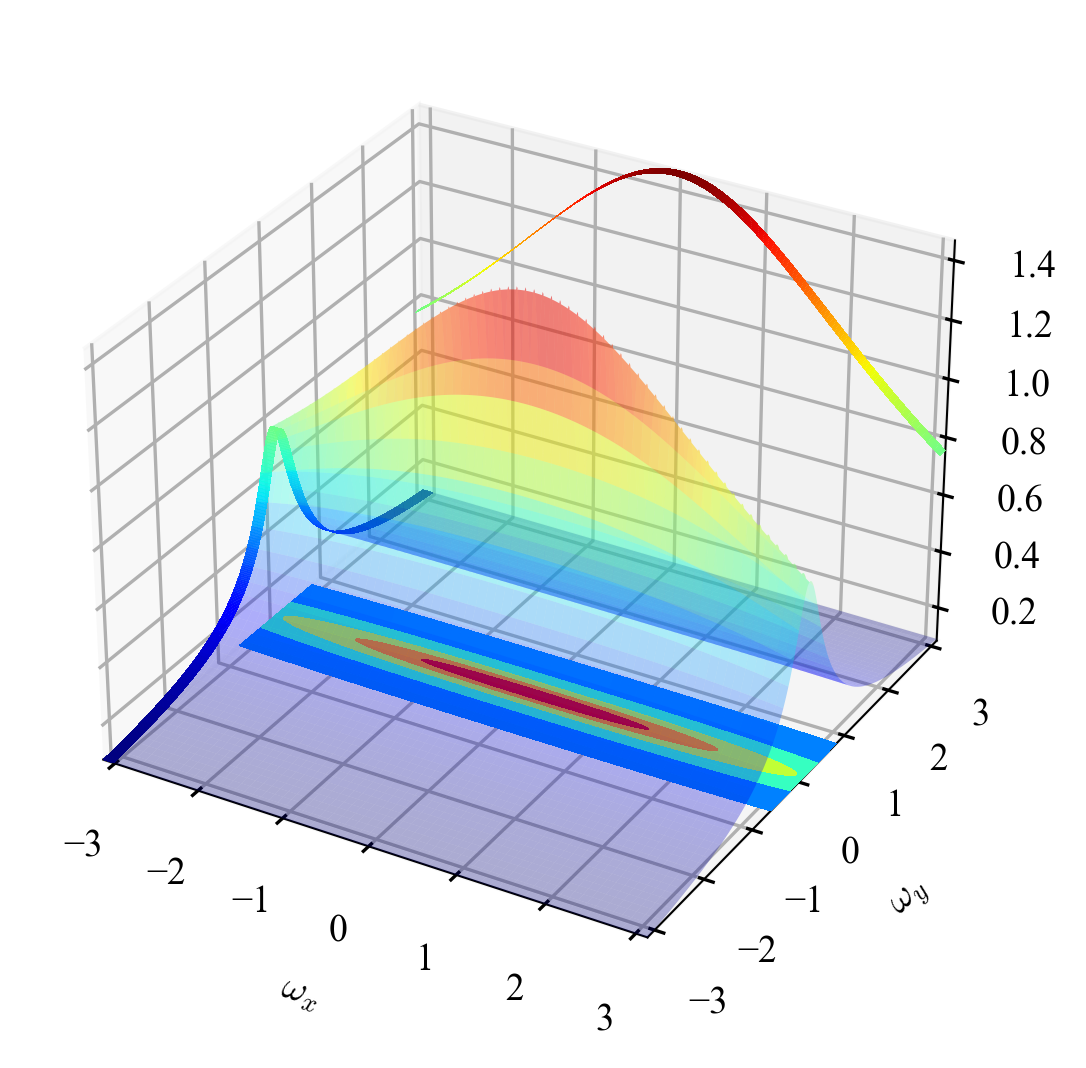}\\[-1pt]
              {\footnotesize Scale: \textcolor{orange}{$\uparrow s_1$}}
            \end{minipage}
            &
            \begin{minipage}{0.16\linewidth}
              \centering
              \includegraphics[width=\linewidth]{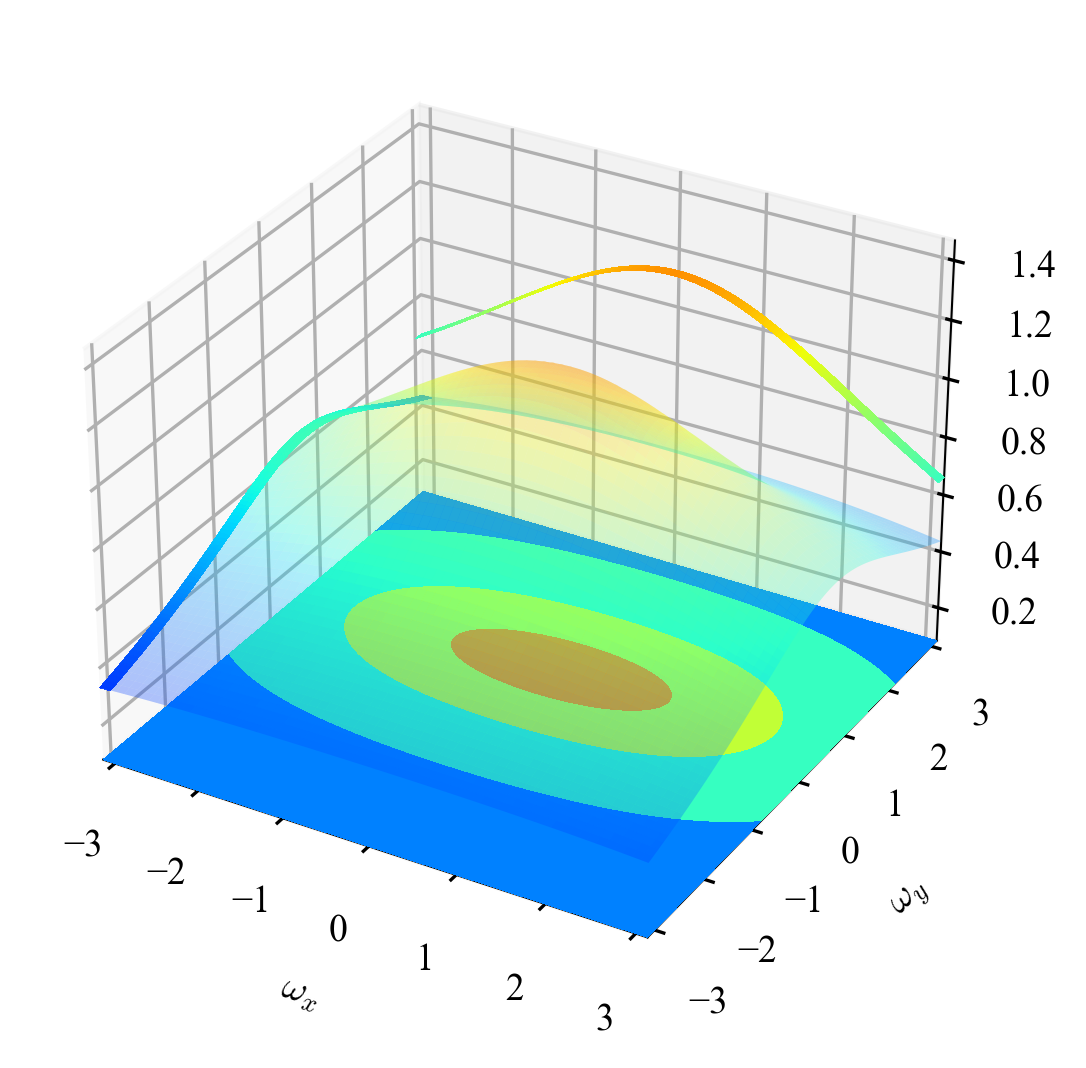}\\[-1pt]
              {\footnotesize $\mathrm{Re}(a_1)$: \textcolor{red}{$\uparrow \alpha_1$}}
            \end{minipage}
            &
            \begin{minipage}{0.16\linewidth}
              \centering
              \includegraphics[width=\linewidth]{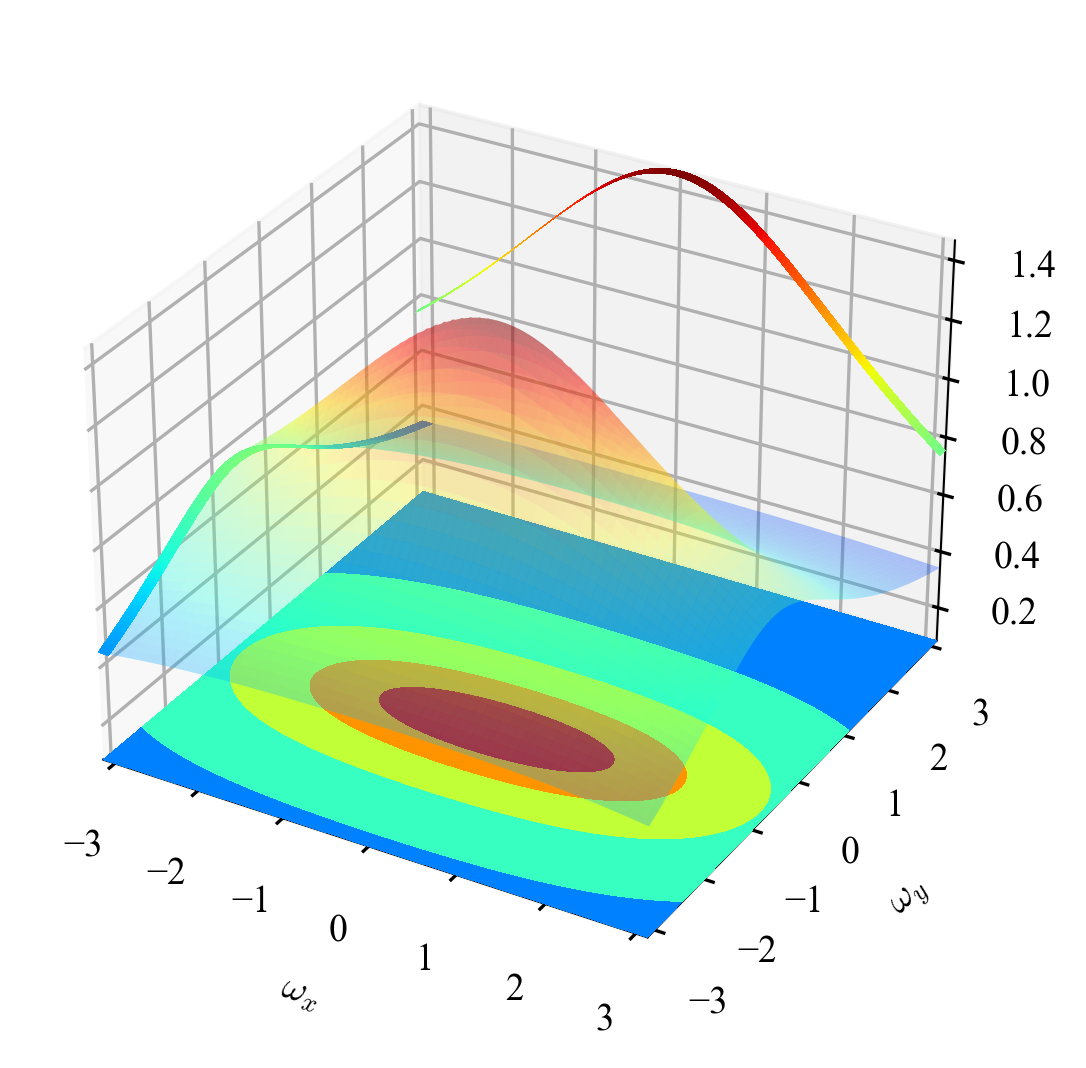}\\[-1pt]
              {\footnotesize $\mathrm{Im}(a_1)$: \textcolor{purple}{$\uparrow \beta_1$}}
            \end{minipage}
            &
            \begin{minipage}{0.16\linewidth}
              \centering
              \includegraphics[width=\linewidth]{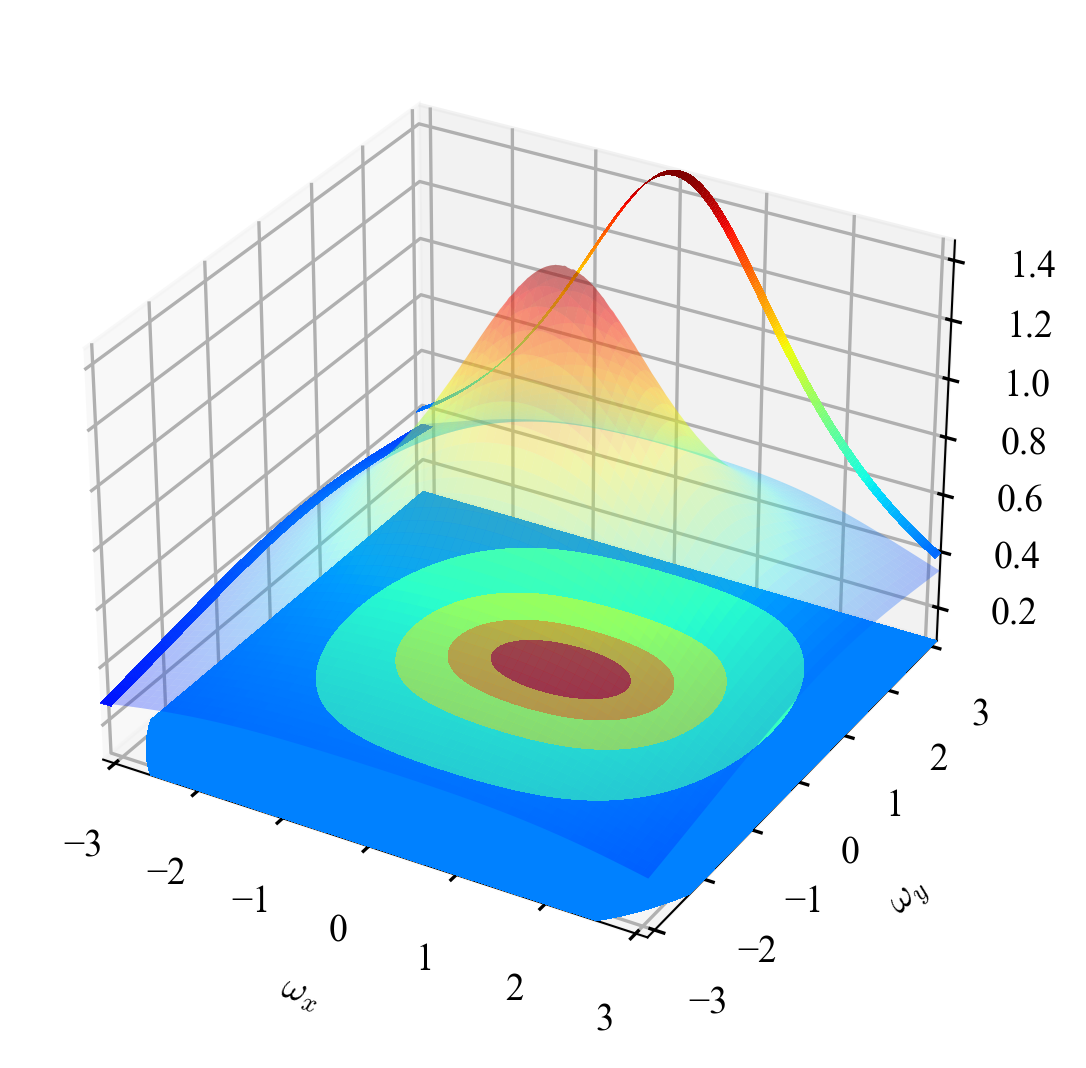}\\[-1pt]
              {\footnotesize Transverse: \textcolor{violet}{$\uparrow \tau_1$}}
            \end{minipage}
          \end{tabular}

          \vspace{4pt}

        \end{minipage}

      \end{minipage}
    };
    \node[anchor=north, font=\bfseries\footnotesize]
      at ([yshift=-2pt]Dcard.north) {Mode Parametrisation};
  \end{tikzpicture}

  \label{fig:sonic_params}
 \caption{Each transfer function $T_m(\omega)$ is parameterized by interpretable 
geometric factors—orientation, scale, complex coefficients, and transverse 
decay—producing a structured family of spectral filters. Shown: parameter 
sweep for mode $T_1$, visualized as $Z=\log\!\left(1+\lvert 
T(\omega_x,\omega_y)\rvert\right)$.}
\end{subfigure}
  \label{fig:sonic_overview_cards}
  \caption{SONIC overview: (a) Residual Block, (b) Spectral Mixing, and (c) Learnable Spectral Modes.}
}

\title{SONIC: Spectral Oriented Neural Invariant Convolutions}

\author{
  Gijs Joppe Moens$^{1,2}$, Regina Beets-Tan$^{1,3}$, Eduardo H. P. Pooch$^{1,3}$\thanks{Code available at \url{https://github.com/GijsMoens/Sonic}} \\
   $^{1}$ Netherlands Cancer Institute, $^{2}$ University of Amsterdam, $^{3}$ Maastricht University \\
  \texttt{\{g.moens, r.beetstan, e.pais.pooch\}@nki.nl}
}

\iclrfinalcopy %
\begin{document}
\maketitle

\begin{figure}[h!]
    \centering
    \resizebox{0.80\textwidth}{!}{
    \input{tikz_receptive_field.tex}
    }
    \label{fig:method_comparison}
\end{figure}

\begin{abstract}
Convolutional Neural Networks (CNNs) rely on fixed-size kernels scanning local patches, which limits their ability to capture global context or long-range dependencies without very deep architectures. Vision Transformers (ViTs), in turn, provide global connectivity but lack spatial inductive bias, depend on explicit positional encodings, and remain tied to the initial patch size. Bridging these limitations requires a representation that is both structured and global.  We introduce \textbf{SONIC (Spectral Oriented Neural Invariant Convolutions)}, a continuous spectral parameterisation that models convolutional operators using a small set of shared, orientation-selective components. These components define smooth responses across the full frequency domain, yielding global receptive fields and filters that adapt naturally across resolutions. Across synthetic benchmarks, large-scale image classification, and 3D medical datasets, SONIC shows improved robustness to geometric transformations, noise, and resolution shifts, and matches or exceeds convolutional, attention-based, and prior spectral architectures with an order of magnitude fewer parameters. These results demonstrate that continuous, orientation-aware spectral parameterisations provide a principled and scalable alternative to conventional spatial and spectral operators.
\end{abstract}

\section{Introduction}

Human visual processing is a remarkably complex and efficient system. It enables us to effortlessly recognise objects, detect and interpret motion, and comprehend complex scenes, adapting seamlessly across varying orientations, scales, resolutions, and even under degraded conditions, where computer vision methods often struggle. Serving as a benchmark due to its exceptional effectiveness under different circumstances, human vision highlights the areas where current artificial systems still exhibit limitations; Bridging this gap remains a central challenge in computer vision, driving the development of models that more closely approximate the versatility and robustness of human perception.\\ \\
Multi-Layer Perceptrons (MLPs), as the earliest neural network models, demonstrated the feasibility of learning complex mappings but lacked the inductive biases required for large-scale vision tasks. Convolutional Neural Networks (CNNs) \citep{lecun2015deep}, widely used for many vision tasks, rely on fixed-size kernels scanning local image patches. While effective for capturing local features like edges and textures, this design limits their ability to understand the overall context or capture long-range dependencies without relying on very deep architectures (as demonstrated by \citet{DBLP:journals/corr/LuoLUZ17}. Critically, their effectiveness is limited by sensitivity to slight geometric variations, including translations (in particular out-of-frame translations), rescalings, rotations, and mild distortions \citep{DBLP:journals/corr/abs-1805-12177}. Vision Transformers (ViTs) \citep{DBLP:journals/corr/abs-2010-11929}, inspired by advances in natural language processing, overcome this limitation by dividing images into sequences of patches and applying self-attention. This design directly models global context and alleviates the locality constraints of CNNs. Nevertheless, the self-attention mechanism is computationally demanding, as its cost grows quadratically with the number of image patches, and thus with the image area, which poses significant challenges for high-resolution inputs. Furthermore, Vision Transformers lack CNN-style spatial inductive biases and therefore require explicit mechanisms (e.g. positional encodings) to model positional relationships, and their accuracy–compute trade-off is closely tied to the chosen patch size. With the proposed method, which enables global receptive fields using significantly fewer parameters, we aim to narrow this conceptual gap and move computer vision models toward resolution-invariant perception, drawing inspiration from the robustness and adaptability of human-like visual processing. 
\paragraph{Contribution} In this paper, we introduce a theoretically grounded spectral framework for multidimensional signals that naturally provides global receptive fields, full convolutional expressiveness, and inherent resolution invariance, offering a lightweight yet versatile foundation that can support progress toward more scalable and adaptable vision models. The remainder of this paper is organised as follows. Section 2 introduces the mathematical preliminaries and related works. Section 3 presents the formulation of the SONIC approach together with implementation details. Section 4 reports the experimental results. Section 5 discusses the limitations of the proposed method and outlines directions for future research.
\section{Background}
Modern vision tasks demand the ability to integrate information over long spatial ranges. Although natural images often exhibit long-range structure, standard convolutions remain bounded by local receptive fields, making standard architectures inefficient, as many layers are effectively used to propagate information across the image rather than to learn increasingly abstract representations.  Across established methods, global context is mostly obtained only indirectly, motivating the study of operators that provide global receptive fields as an intrinsic property of a single layer.  This section reviews the mathematical foundations of such operators and develops the framework of spectral operators that underpins our approach. 
\paragraph{Spatial-domain operators.}
In the spatial domain, enlarging the receptive field requires expanding the support of the discrete kernel. Large-kernel convolutions increase the neighbourhood size directly \citep{ding2022scalingkernels31x31revisiting}, dilated convolutions introduce gaps to cover larger regions without increasing the number of parameters \citep{yu2015dilatedconvolutions} and non-local \citep{wang2018nonlocalneuralnetworks} methods target long-range interactions; however, despite their effectiveness, convolutional layers implement filtering over a fixed sampling grid, an approach that implicitly assumes locality and smooth variation in the underlying signal. These assumptions hold for small neighborhoods but break down over large spatial ranges, where long-range structure cannot be captured efficiently through local interactions alone. As receptive fields expand, spatial filters become increasingly tied to the image resolution and scale, limiting their generalization and efficiency. Another well-studied strategy is to use self-attention mechanisms \citep{chen2018a2netsdoubleattentionnetworks,dosovitskiy2021imageworth16x16words}, which compute pairwise interactions across all spatial positions and therefore provide a principled way to model long-range dependencies. However, these approaches incur computational and memory costs that grow rapidly with image resolution: large kernels scale with their area $O(K^2)$, attention scales quadratically with the number of tokens, and as resolution increases, these scaling properties make such mechanisms difficult to deploy efficiently, especially in high-resolution domains where global context is important but computational budgets are constrained.
\begin{figure}[H]
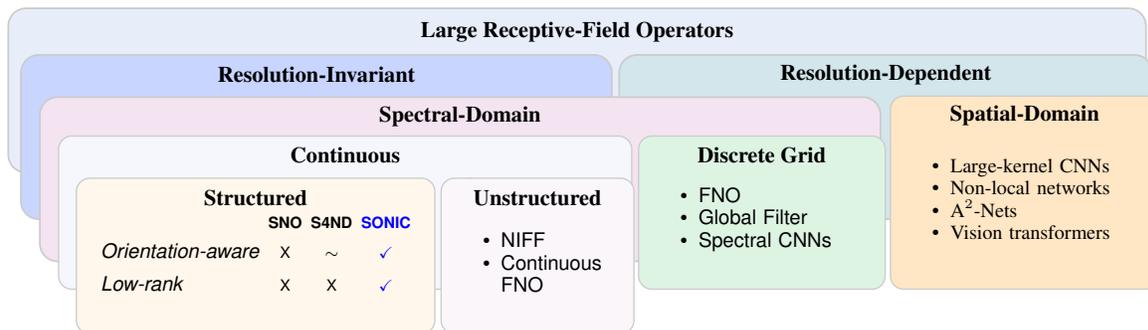

\centering
\resizebox{0.9\linewidth}{!}{%
  \LRFOTaxonomy
}
\caption{(a) Taxonomy of large receptive-field operators.}
\label{fig:taxonomy}
\end{figure}

\paragraph{Spectral-domain operators.}
An alternative paradigm achieves global receptive fields by representing operators directly in the 
frequency domain \citep{rippel2015spectralrepresentationsconvolutionalneural}.  This approach uses the fact that every linear, shift-invariant operator on $\mathbb{R}^D$ is fully 
characterised, enabling information to 
propagate globally through a frequency-wise multiplication.

Let $D\in\mathbb{N}$ and consider vector-valued signals
\[
x : \mathbb{R}^D \to \mathbb{C}^{C_{\mathrm{in}}}, \qquad 
y : \mathbb{R}^D \to \mathbb{C}^{C_{\mathrm{out}}}.
\]
For a sufficiently regular scalar function $f : \mathbb{R}^D \to \mathbb{C}$, the Fourier transform is
\begin{equation}
\mathcal{F}_D[f](\boldsymbol{\omega})
= \int_{\mathbb{R}^D} f(\mathbf{x})\,e^{-i\,\boldsymbol{\omega}\cdot\mathbf{x}}\,d\mathbf{x},
    \qquad \boldsymbol{\omega}\in\mathbb{R}^D.
\end{equation}
and extends component-wise to vector-valued functions.  
A linear, shift-invariant operator acting on $x$ has a convolution representation:
\begin{equation}
y(\mathbf{x}) = \int_{\mathbb{R}^D} k(\mathbf{x}-\mathbf{z})\, x(\mathbf{z}) \, d\mathbf{z},
\end{equation}
where $k : \mathbb{R}^D \to \mathbb{C}^{C_{\mathrm{out}}\times C_{\mathrm{in}}}$.  
The Convolution Theorem gives
\begin{equation}
\mathcal{F}_D[y](\boldsymbol{\omega})
= \widehat{k}(\boldsymbol{\omega})\,\mathcal{F}_D[x](\boldsymbol{\omega}),
\qquad \boldsymbol{\omega}\in\mathbb{R}^D.
\end{equation}
where $\widehat{k}(\boldsymbol{\omega})$ is the Fourier transform of a spatial kernel in the
classical convolution setting.  
In spectral neural methods, however, we do not constrain the operator to arise from any
finite-support spatial kernel.  
Instead, we define the spectral kernel directly by
\begin{equation}
\widehat{k}(\boldsymbol{\omega}) := \widehat{H}(\boldsymbol{\omega}),
\end{equation}
where $\widehat{H}(\boldsymbol{\omega})$ is the learnable frequency response of the operator. This viewpoint treats $\widehat{H}$ as the primary parametrisation, enabling general global and resolution-invariant operators beyond those that correspond to discrete spatial kernels. 
In practice, images are sampled on a discrete grid.
Let the spatial domain be discretised using $N_1,\dots,N_D$ samples along each axis, with pixel spacings $\Delta_1,\dots,\Delta_D$.
The corresponding DFT frequency sets are given by
\begin{equation}
\Omega_d
    = 2\pi \left\{
        \frac{k_d}{N_d\,\Delta_d}
        \;\middle|\;
        k_d = -\left\lfloor\tfrac{N_d}{2}\right\rfloor,\dots,
              \left\lceil\tfrac{N_d}{2}\right\rceil - 1
      \right\},
\quad d=1,\dots,D.
\end{equation}
The full frequency grid is the Cartesian product $\Omega = \Omega_{1} \times \cdots \times \Omega_{D}$, containing $N=N_1\cdots N_D$ discrete frequencies. The DFT samples $\widehat{x}$ are defined at frequencies $\boldsymbol{\omega}_n \in \Omega$.

\paragraph{Resolution invariance} We formalise resolution invariance by defining the operator via a continuous spectral symbol that is independent of the sampling grid. Let
\begin{equation}
\widehat{H}_{\theta} : \mathbb{R}^D \to \mathbb{C}^{C_{\mathrm{out}}\times C_{\mathrm{in}}}
\end{equation}
be a continuous function parameterised by $\theta$. Given a discretisation $(N,\Delta)$ with Fourier grid $\Omega_{N,\Delta}$, the discretised operator is obtained via sampling:
\begin{equation}
\widehat{y}^{(N,\Delta)}(\boldsymbol{\omega}_n)
    = \widehat{H}_{\theta}(\boldsymbol{\omega}_n)\,
      \widehat{x}^{(N,\Delta)}(\boldsymbol{\omega}_n), 
    \qquad \boldsymbol{\omega}_n \in \Omega_{N,\Delta}.
\end{equation}
We term the operator resolution-invariant if $\theta$ depends only on the underlying physics of the layer, not on the discretisation $(N,\Delta)$. Changing resolution then simply corresponds to resampling the same continuous function $\widehat{H}_{\theta}$ on a new grid. GFNet \citep{rao2021globalfilternetworksimage} and FNO \citep{li2021fourierneuraloperatorparametric} parameterise $\widehat{H}$ directly on the discrete FFT grid:  
GFNet learns a complex mask of size $N$, and FNO learns a fixed number of low-frequency 
coefficients.  
Since these coefficients correspond to specific frequency indices, changing resolution alters the 
operator itself.  Thus, such models do not define a true resolution-invariant convolution operator.
\paragraph{Continuous Spectral Operators}
A principled way to overcome this limitation is to define the operator directly in the continuous Fourier domain and then evaluate it on the discrete grid provided by the data.  In this formulation, the spectral symbol $\widehat{H}(\boldsymbol{\omega})$ is a continuous function of frequency, independent of the sampling pattern, and the discrete operator is obtained merely by sampling $\widehat{H}$ at the DFT frequencies of the current resolution.  This yields a truly resolution-invariant parameterisation with global receptive fields. Two families of such continuous spectral operators appear in the literature:
\begin{itemize}
\item \textbf{Unstructured continuous operators}, which learn 
$\widehat{H}(\boldsymbol{\omega})$ as a general continuous function of frequency, typically 
via a small MLP or other low-dimensional parametrisation.  
Such models include neural implicit spectral filters and continuous FNO variants \citep{grabinski2024largegetslearninginfinitely, kabri2023resolutioninvariantimageclassificationbased}, where the network outputs a complex response for any $\boldsymbol{\omega}\in\mathbb{R}^D$.  While this provides maximal flexibility and full continuity in $\boldsymbol{\omega}$, these parameterisations are usually isotropic or weakly anisotropic: the learned response depends primarily on $\|\boldsymbol{\omega}\|$ unless orientation structure is encoded explicitly.  Moreover, the channel mixing remains fully dense, offering little inductive bias regarding frequency orientation or cross-channel structure.  As a result, these operators can be expressive but often parameter-inefficient. 
\item \textbf{Structured continuous operators} imposes additional structure on $\widehat{H}(\boldsymbol{\omega})$ through basis expansions, separability assumptions, or functional templates. Examples include SNO \citep{fanaskov2024spectralneuraloperators}, which expands the symbol in a fixed orthogonal basis, and multidimensional SSM-based kernels like S4ND \citep{nguyen2022s4ndmodelingimagesvideos}, which impose axis-aligned or separable constructions inherited from one-dimensional state-space models. Although these multidimensional SSMs are not typically framed as spectral operators, their learned kernels do admit a structured frequency-domain representation and can be interpreted through the same lens as spectral methods. For completeness, we provide the frequency-domain form of S4ND in the appendix.
 These approaches provide improved inductive bias and parameter efficiency by coupling nearby frequencies and reusing spectral modes across channels. While more efficient, most structured models remain tied to coordinate axes. Their separability limits their ability to capture oriented or anisotropic patterns that lie along 
general directions in frequency space.\end{itemize}

Natural images contain oriented structures such as edges, textures, and oscillations that 
correspond to directional features in frequency space.  
Standard separable parameterisations cannot easily model such behaviour.  
To represent interactions along arbitrary frequency directions, one must go beyond axis-aligned 
or tensor-product constructions and design spectral operators whose modes explicitly encode 
orientation in $\mathbb{R}^D$.\\ \\
In this paper, we address this limitation by introducing SONIC: Spectral Oriented Neural Invariant Convolutions. SONIC is a Structured Continuous Spectral Operator that moves beyond axis-aligned constructions by explicitly parameterizing the spectral symbol $\widehat{H}_{\theta}(\boldsymbol{\omega})$ as a superposition of directional modes. This allows us to learn complex, oriented features in the frequency domain that are fully resolution-invariant, parameter-efficient, and inherently adapted to capturing the anisotropic structures present in natural signals. 

\section{Method}
\label{sec:method}
\paragraph{Overview.}
Many spectral neural methods are either axis–separable (efficient but limited in orientation) or fully nonlocal (powerful but inefficient and not spectrally faithful). Starting from linear time-invariant (LTI) systems, we extend the formulation to $N$-dimensional signals in the frequency domain, yielding a compact spectral representation. This framework models linear, shift–invariant operators through a shared low–rank structure, where oriented spectral transfer functions are applied at each frequency and mixed across channels by learned matrices $B$ and $C$.
\subsection{Formulation}
Our method draws inspiration from the analytic structure of linear time-invariant systems, the same foundation underlying modern state-space models. To make this connection precise, consider the continuous-time LTI state-space system:
\begin{equation}
\dot{\mathbf{z}}(t) = \mathbf{A}\mathbf{x}(t) + \mathbf{B}\mathbf{u}(t), 
\qquad 
\mathbf{y}(t) = \mathbf{C}\mathbf{x}(t),
\label{eq:lti_state}\end{equation}
with zero initial condition.  
Its impulse response is obtained by setting $\mathbf{u}(t)=\delta(t)$:
\begin{equation}
\mathbf{K}(t) \;=\; \mathbf{C}\,e^{\mathbf{A}t}\,\mathbf{B}, 
\qquad t \ge 0.
\label{eq:lti_impulse}
\end{equation}
The output equals the convolution of the input with the impulse response:
\begin{equation}
(\mathbf{K} * \mathbf{u})(t)
\;=\; \int_{0}^{\infty} \mathbf{C}\,e^{\mathbf{A}\tau}\,\mathbf{B}\;
\mathbf{u}(t-\tau)\,d\tau.
\label{eq:lti_conv}
\end{equation}
Taking the Laplace transform of the impulse response 
(derivations provided in  Appendix~\hyperref[appendix:convolution_theorem]{C}),
\begin{equation}
H(s)
\;=\;
\mathcal{L}\{ \mathbf{K}(t) \}(s)
=\;
\mathbf{C}(s\mathbf{I}-\mathbf{A})^{-1}\mathbf{B}.
\label{eq:transfer}
\end{equation}
The expression $H(s)$ above is the standard resolvent form that
characterises the frequency response of a stable linear
time–invariant system.  We use this structure only as a modelling
template: by replacing the scalar Laplace variable $s$ with the
multi-dimensional spatial frequency $\boldsymbol{\omega}$, we obtain an
analytic spectral parameterisation that inherits the smooth and
structured behaviour of resolvent filters in $D$ dimensions.

Let the input be $x \in \mathbb{R}^{C \times N_1 \times \cdots
\times N_D}$ and the output $y \in \mathbb{R}^{K \times N_1 \times
\cdots \times N_D}$. We denote their $D$-dimensional discrete Fourier
transforms by
\[
\widehat{x} = \mathcal{F}_D[x],
\qquad
y = \mathcal{F}_D^{-1}[\widehat{y}].
\]

Central to our method is the transfer function
$T_m(\boldsymbol{\omega})$, which defines the frequency response of a
single mode. For each mode $m=1,\dots,M$ we set
\begin{equation}
\label{eq:mode}
T_m(\boldsymbol{\omega})
\;=\;
\frac{1}{\, i\, s_m \, (\boldsymbol{\omega}\!\cdot\!\boldsymbol{v_m})
\;-\; a_m
\;+\; \tau_m \,\|
(I - \boldsymbol{v_m v_m^\top})\boldsymbol{\omega}
\|_2^{2}},
\end{equation}
Where each mode is parameterised by:
(1) the orientation $\boldsymbol{v_m}\in\mathbb{R}^D$ with
$\|\boldsymbol{v_m}\|_2=1$;  
(2) the scale $s_m>0$ controlling spectral selectivity;  
(3) the real part $\operatorname{Re}(a_m)$ introducing damping;  
(4) the imaginary part $\operatorname{Im}(a_m)$ governing oscillatory behaviour; and  
(5) the transverse penalty $\tau_m\ge 0$ controlling decay orthogonal to
$\boldsymbol{v_m}$.  
Together, these parameters shape the amplitude, orientation, and
oscillatory nature of each spectral mode. The denominator replicates the resolvent structure of an LTI system. SONIC adopts this template by substituting the Laplace variable $s$ with the oriented 
frequency component $i\,s_m(\boldsymbol{\omega}\ \cdot \boldsymbol{v}_m)$ and by adding 
a transverse decay term that enforces anisotropic filtering.\\ \\
Rather than learning an unconstrained response
$\widehat{\mathbf{H}}(\boldsymbol{\omega})$ for every frequency,
SONIC factorises the spectral operator through $M$ shared modes with entrywise form:
\begin{equation}
\label{eq:rankM}
\widehat{H}_{k,c}(\boldsymbol{\omega})
=
\sum_{m=1}^M C_{km}\,T_m(\boldsymbol{\omega})\,B_{mc}.
\end{equation}

\noindent
where $B \in \mathbb{C}^{M \times C}$ and $C \in \mathbb{C}^{K \times M}$. Given this factorised spectral response, the frequency-wise filtering
applied to the input DFT is
\begin{equation}
\label{eq:mimo}
\widehat{y}_k(\boldsymbol{\omega})
=
\sum_{c=1}^{C}
\widehat{H}_{k,c}(\boldsymbol{\omega})\,\widehat{x}_c(\boldsymbol{\omega}),
\qquad
k=1,\dots,K,\;\boldsymbol{\omega}\in\Omega,
\end{equation}
where $\widehat{H}_{k,c}(\boldsymbol{\omega})$ is the frequency response of
the $(c\!\to\!k)$ channel filter. This decomposition yields a compact, low-rank representation of the
spectral operator, enabling expressive but parameter-efficient filtering. Following the frequency-domain filtering, the spatial output is added to a learnable skip projection and then passed through a pointwise nonlinearity, yielding the next-layer activation $x^{(\ell+1)}$:
\begin{equation}
x^{(\ell+1)} = \sigma\!\big(y^{(\ell)} + W_sx^{(\ell)}\big).
\end{equation}
This nonlinear recursion allows multiple SONIC blocks to be stacked, providing depth-wise expressivity in the same manner as conventional convolutional architectures.
Although SONIC is not a state-space model, its mode parameterisation is inspired by resolvent structures of linear time-invariant systems.
Appendix  \hyperref[appendix:convolution_theorem]{C} shows that restricting SONIC’s
orientations to the coordinate axes yields the Multidimensional SSM form.

\subsection{Intuition}
We use a compact collection of oriented modes that are shared across channels. Instead of learning an unconstrained spectrum for every input–output pair, each mode has a learnable analytic shape with a few learnable knobs, yielding interpretable, spatially localised filters. We also illustrate the effect of each parameter in Figure  ~\hyperref[fig:sonic_overview_cards]{2}.

Each mode learns a preferred direction via a unit vector $v_m$, a compass needle in frequency space. Any frequency vector $\boldsymbol{\omega}$ decomposes uniquely into components along and across to this needle:
\[
\omega_{|| m} := \boldsymbol{\omega}\!\cdot\! \boldsymbol{v_m} \quad, \qquad
\boldsymbol{\omega}_{\perp m} := (I - \boldsymbol{v_m v_m^\top})\boldsymbol{\omega} \quad.
\]
 The mode passes slow variation along its needle and increasingly damps faster oscillations in that direction, so gently varying, needle-aligned content is emphasized while rapidly oscillating content along the axis is attenuated. It also suppresses energy that lies across the needle, so components that are not aligned with the needle’s orientation contribute less. In spatial terms, the resulting kernel is stretched along $v_m$ (making it sensitive to lines, flows, or ridges in that direction) and compressed across it.\\ \\ The scale parameter $s_m$ regulates the mode’s spectral selectivity. Small values produce a broad response that pools over a wide band of along-axis frequencies, acting as an orientation-aware smoother that preserves coarse structure while suppressing fine fluctuations. Large values narrow the passband and sharpen selectivity, emphasizing only a thin slice of along-axis variation; in the spatial domain, this corresponds to a longer, more finely structured kernel along $v_m$. During learning, $s_m$ adapts locally to the content of the signal: scenes dominated by broad shapes tend to drive $s_m$ down, while scenes rich in fine oriented detail push it up.

By contrast, the complex coefficient $a_m$ governs the global dynamics of each mode. Its real part controls damping, ensuring stability, while its imaginary part introduces oscillations that can be amplified or suppressed. These oscillations enrich the representation, allowing the mode to capture structured patterns in the plane. Unlike $s_m$, which tunes frequency 
selectivity along the axis, $a_m$ balances between 
smoothness and oscillatory structure: smoother, slowly varying signals encourage stronger damping and broader low-pass behavior, whereas signals with repetitive, oriented fine-scale structure favor a smaller imaginary component that preserves such fine patterns.

Finally, the transverse penalty $\tau_m \ge 0$ pushes down frequencies that point away from $v_m$. This sharpens directional selectivity by suppressing leakage into neighboring directions and, in higher dimensions, prevents degenerate, plane-like responses. Intuitively, larger $\tau_m$ clamps the response tightly around the chosen axis, whereas smaller $\tau_m$ allows more lateral spread. 

\begin{figure}[H]
  \centering
  \resizebox{0.9\linewidth}{!}{%
    \begin{minipage}{\linewidth}
      \AllFigs
    \end{minipage}
  }
\end{figure}
Conceptually, the modes, after the spectral transfer, form a small dictionary of directional behaviors, while separate learned mixing weights decide how each input channel contributes to, and each output channel draws from, the same dictionary. This keeps parameters modest and encourages reuse of structure across channels.
After building the modes, we let the model mix the different modes by $\mathbf{C}$ and $\mathbf{B}$, this ensures that each channel c mapping to output k can be a unique superposition of all constructed modes. The key distinction compared to other spectral methods is the parameterisation of the spectral domain. Conventional spectral neural operators employ an unconstrained, discrete representation, assigning independent complex coefficients to each sampled frequency $\omega_k$. In contrast, SONIC utilises a structured low-rank factorisation built from a small set of shared spectral modes. Each mode is governed by a smooth, orientation-sensitive transfer function $T_m(\omega)$, yielding a continuous and anisotropic dependence on the frequency variable. This induces substantial parameter sharing across both frequencies and channels, in contrast to traditional spectral approaches, whose representations are frequency-wise independent and lack functional coherence in $ \omega$. 
\paragraph{Resolution Invariance} Crucially, all of these filters are parameterized directly in the continuous spectral domain. This means their definition does not depend on the size or sampling rate of the image: defining filters as continuous functions of $\boldsymbol{\omega}$ decouples them from any particular grid size or sampling rate; the same response formula is evaluated on whatever DFT grid the data induces, yielding a resolution-invariant filter. This distinguishes our approach from spatial-domain kernels, whose size and shape are tied to a fixed grid.  We made some minor adjustments to ensure resolution invariance:
To make the directional parameters resolution invariant, we express directions in physical units and normalise:
\begin{equation}
D_\Delta \;=\; \mathrm{diag}(\Delta_1,\ldots,\Delta_D),\qquad
\boldsymbol{\tilde v_m} \;=\; D_\Delta^{-1} \boldsymbol{v_m},\qquad
\hat v_m \;=\; \frac{\boldsymbol{\tilde v_m}}{\|\boldsymbol{\tilde v_m}\|_2}\, .
\end{equation}
This resolution-aware formulation can be exploited during training, as also proposed in \citet{nguyen2022s4ndmodelingimagesvideos}. Beyond efficiency, it is particularly relevant in domains where resolution dependence is intrinsic, such as medical imaging, remote sensing, and microscopy.
\paragraph{Computation} The number of learnable real scalars is:
\[
\underbrace{2KM}_{C^{\text{re}},C^{im}}
\;+\;
\underbrace{2MC}_{B^{re},B^{im}}
\;+\;
\underbrace{(4+D)M+1}_{a^{\mathrm{re}},\,a^{\mathrm{im}}, \,s,\,v,\tau \ \in\mathbb{R}^2},
\]
For the FFT transformation we used the highly optimized VkFFT library \citep{10036080}, with per-transform cost \(O(N\log N)\) for a single (complex) channel. The spectral forward pass performs one DFT per input channel and one inverse DFT per output channel, plus $O(M(C{+}K))$ complex multiplications per frequency. The forward pass consists of one DFT per input channel and one inverse DFT per output channel, s
with cost
\[
O(CN\log N) \quad \text{and} \quad O(KN\log N),
\]
where $N = \prod_{d=1}^{D} N_d$ is the total number of spatial points.
In addition, frequency-wise multiplications incur a cost of
\[
O\!\big(M(C{+}K)N\big),
\]
since each of the $M$ modes couples inputs and outputs across all frequencies.
The total complexity is therefore
\[
O\!\big((C{+}K)N\log N \;+\; M(C{+}K)N\big).
\]
SONIC is thus particularly attractive for large receptive fields (where $d$ is large or even global), since the cost remains manageable and the parameter count remains compact.
\section{Empirical validation}
\paragraph{\textbf{SynthShape}}
To evaluate the sensitivity of models to geometric variations, we introduce SynthShape (Synthetic Shape Dataset), a simple 64x64 synthetic geometric shape–based segmentation benchmark. We assess model generalization by measuring performance under controlled perturbations, Rescaling (including interpolation artefacts), in-plane rotation, out-of-frame translation, geometric distortion, and additive Gaussian noise. All experiments are conducted using 5-fold cross-validation. Furthermore, we introduce \textbf{HalliGalli}, a controlled spatial-reasoning task modeled after the well-known game, designed to test effective long-range dependency modelling rather than theoretical receptive field size. The task is to classify a central patch according to whether exactly two matching shapes appear in the four distant corners; the centre itself carries no class signal. Since the task depends on structure that cannot be captured within any local receptive field, purely local models fail. Architectures with untargeted global filters either fail due to missing orientation or degrade under Gaussian noise, as their large receptive fields accumulate noise over a broad spatial region. SONIC successfully solves the HalliGalli task and remains robust under inference-time noise, demonstrating the effectiveness of its globally oriented structured receptive field.Further implementation details are provided in Appendix \ref{appendix:implementation_gs}. 
\begin{table*}[t]
\caption{Comparison of ConvNet, ViT, S4ND, NIFF, GFNet, and SonicNet performance on SynthShape under geometric variations (left), and qualitative examples from SynthShape and HalliGalli-SRT (right).}
\label{tab:geosynth_full_results}
\centering

\begin{minipage}[t]{0.55\textwidth}
    \vspace{5pt} %
    \renewcommand{\arraystretch}{1.25}
    \resizebox{\textwidth}{!}{%
   \begin{tabular}{llcccccc}
    \toprule
    \textbf{Experiment} & \textbf{Value} 
    & \textbf{ConvNet} & \textbf{ViT} & \textbf{S4ND} & \textbf{NIFF} & \textbf{GFNet} & \textbf{SonicNet} \\
    \midrule
    \multicolumn{2}{l}{\textbf{Parameter count (M)}} 
    \scanrow{0.153, 0.471, 0.186, 0.042, 0.415, 0.072}
    & \perfcellrev{0.153} & \perfcellrev{0.468} & \perfcellrev{0.186}
    & \perfcellrev{0.042} & \perfcellrev{0.415} & \perfcellrev{0.072} \\
    \multicolumn{2}{l}{\textbf{GMACs}}               
    \scanrow{0.156, 0.012, 0.023, 0.041, 0.139, 0.006}
    & \perfcell{0.156} & \perfcell{0.012} & \perfcell{0.023} & \perfcell{0.041} & \perfcell{0.139} & \perfcell{0.006} \\
    \midrule
    \midrule

    \textbf{Distortion} 
    & 2.0 
    \scanrow{0.97, 0.88, 0.55, 0.94, 0.66, 0.97}
    & \perfcell{0.97} & \perfcell{0.88} & \perfcell{0.84} & \perfcell{0.94} & \perfcell{0.66} & \perfcell{0.97} \\
    & 4.0  
    \scanrow{0.96, 0.88, 0.55, 0.94, 0.62, 0.96} 
    & \perfcell{0.96} & \perfcell{0.88} & \perfcell{0.83} & \perfcell{0.94} & \perfcell{0.62} & \perfcell{0.96} \\
    & 6.0  
    \scanrow{0.94, 0.87, 0.54, 0.92, 0.60, 0.96}
    & \perfcell{0.94} & \perfcell{0.87} & \perfcell{0.82} & \perfcell{0.92} & \perfcell{0.60} & \perfcell{0.96} \\
    \midrule

    \textbf{Gaussian Noise ($\sigma$)} 
    & 0.1 
    \scanrow{0.98, 0.78, 0.52, 0.99, 0.73, 0.98}
    & \perfcell{0.98} & \perfcell{0.78} & \perfcell{0.89} & \perfcell{0.99} & \perfcell{0.73} & \perfcell{0.98} \\
    & 0.2 
    \scanrow{0.85, 0.44, 0.33, 0.92, 0.31, 0.79}
    & \perfcell{0.85} & \perfcell{0.44} & \perfcell{0.67} & \perfcell{0.92} & \perfcell{0.31} & \perfcell{0.79} \\
    & 0.3 
    \scanrow{0.58, 0.32, 0.26, 0.71, 0.19, 0.60}
    & \perfcell{0.58} & \perfcell{0.32} & \perfcell{0.43} & \perfcell{0.71} & \perfcell{0.19} & \perfcell{0.60} \\
    \midrule

    \textbf{Rescaling} 
    & 0.75 
    \scanrow{0.84, 0.73, 0.40, 0.78, 0.44, 0.86}
    & \perfcell{0.84} & \perfcell{0.73} & \perfcell{0.49} & \perfcell{0.78} & \perfcell{0.44} & \perfcell{0.86} \\
    & 1.00$^{*}$ 
    \scanrow{0.99, 0.94, 0.59, 1.00, 0.92, 1.00}
    & \perfcell{0.99} & \perfcell{0.94} & \perfcell{0.93} & \perfcell{1.00} & \perfcell{0.92} & \perfcell{1.00} \\
    & 1.50 
    \scanrow{0.62, 0.68, 0.36, 0.59, 0.37, 0.74}
    & \perfcell{0.62} & \perfcell{0.68} & \perfcell{0.32} & \perfcell{0.59} & \perfcell{0.37} & \perfcell{0.74} \\
    \midrule

    \textbf{Rotation ($^\circ$)} 
    & 15 
    \scanrow{0.69, 0.66, 0.46, 0.70, 0.44, 0.75}
    & \perfcell{0.69} & \perfcell{0.66} & \perfcell{0.68} & \perfcell{0.70} & \perfcell{0.44} & \perfcell{0.75} \\
    & 30 
    \scanrow{0.28, 0.32, 0.38, 0.30, 0.30, 0.23}
    & \perfcell{0.28} & \perfcell{0.32} & \perfcell{0.50} & \perfcell{0.30} & \perfcell{0.30} & \perfcell{0.23} \\
    & 45 
    \scanrow{0.28, 0.30, 0.37, 0.29, 0.28, 0.24}
    & \perfcell{0.28} & \perfcell{0.30} & \perfcell{0.44} & \perfcell{0.29} & \perfcell{0.28} & \perfcell{0.24} \\
    \midrule

    \textbf{Translation (\%)} 
    & 10 
    \scanrow{0.92, 0.87, 0.54, 0.89, 0.53, 0.95}
    & \perfcell{0.92} & \perfcell{0.87} & \perfcell{0.76} & \perfcell{0.89} & \perfcell{0.53} & \perfcell{0.95} \\
    & 20 
    \scanrow{0.96, 0.90, 0.57, 0.96, 0.74, 0.97}
    & \perfcell{0.96} & \perfcell{0.90} & \perfcell{0.76} & \perfcell{0.96} & \perfcell{0.74} & \perfcell{0.97} \\
    & 30 
    \scanrow{0.91, 0.88, 0.52, 0.88, 0.56, 0.93}
    & \perfcell{0.91} & \perfcell{0.88} & \perfcell{0.77} & \perfcell{0.88} & \perfcell{0.56} & \perfcell{0.93} \\
    \midrule

    \textbf{Combined} 
    & 10 
    \scanrow{0.85, 0.76, 0.48, 0.71, 0.36, 0.90}
    & \perfcell{0.85} & \perfcell{0.76} & \perfcell{0.47} & \perfcell{0.71} & \perfcell{0.36} & \perfcell{0.90} \\
    & 20 
    \scanrow{0.62, 0.59, 0.39, 0.50, 0.28, 0.76}
    & \perfcell{0.62} & \perfcell{0.59} & \perfcell{0.30} & \perfcell{0.50} & \perfcell{0.28} & \perfcell{0.76} \\
    & 30 
    \scanrow{0.41, 0.37, 0.32, 0.40, 0.25, 0.48}
    & \perfcell{0.41} & \perfcell{0.37} & \perfcell{0.24} & \perfcell{0.40} & \perfcell{0.25} & \perfcell{0.48} \\
    \midrule

     \textbf{HalliGalli} 
    & & 0.42 & 0.33 & 0.62 & 1.00 & 0.71 & 1.00 \\
    \textbf{HalliGalli ($\sigma=0.1$)} 
    & & 0.33 & 0.33 & 0.49 & 0.56 & 0.37 & 0.86 \\
    \bottomrule
    \multicolumn{8}{l}{$^{*}$ Validation accuracy on the training task.} \\
\end{tabular}
    }
\end{minipage}
\hfill
\begin{minipage}[t]{0.4\textwidth}
    \vspace{3pt} %
    \centering
    \textbf{SynthShape}\\[0.3em]
    \hspace{3mm}\includegraphics[width=.95\linewidth]{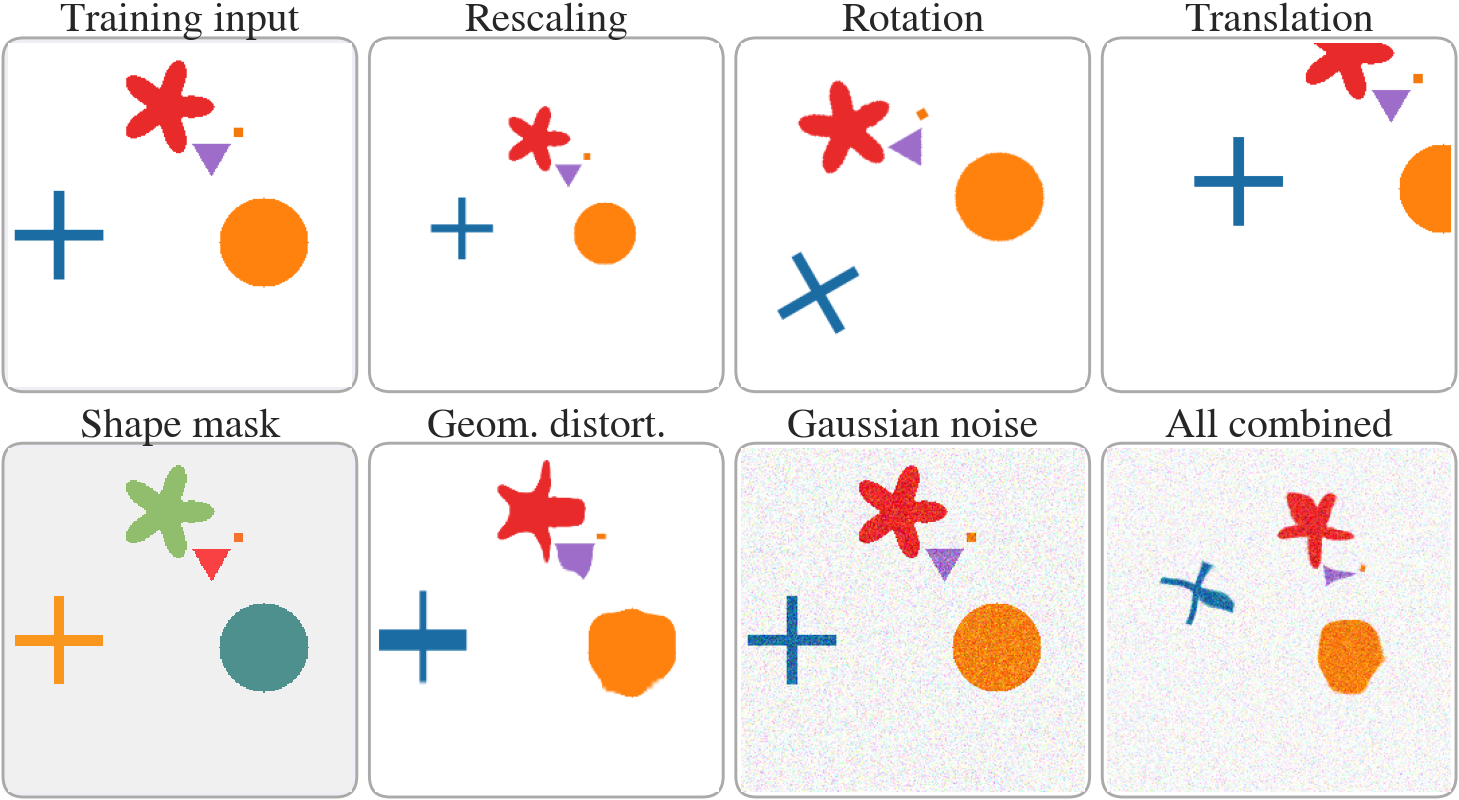}\\[0.4em]
    \vspace{10pt}
    
    \textbf{HalliGalli}\\[0.3em]
    \includegraphics[width=\linewidth]{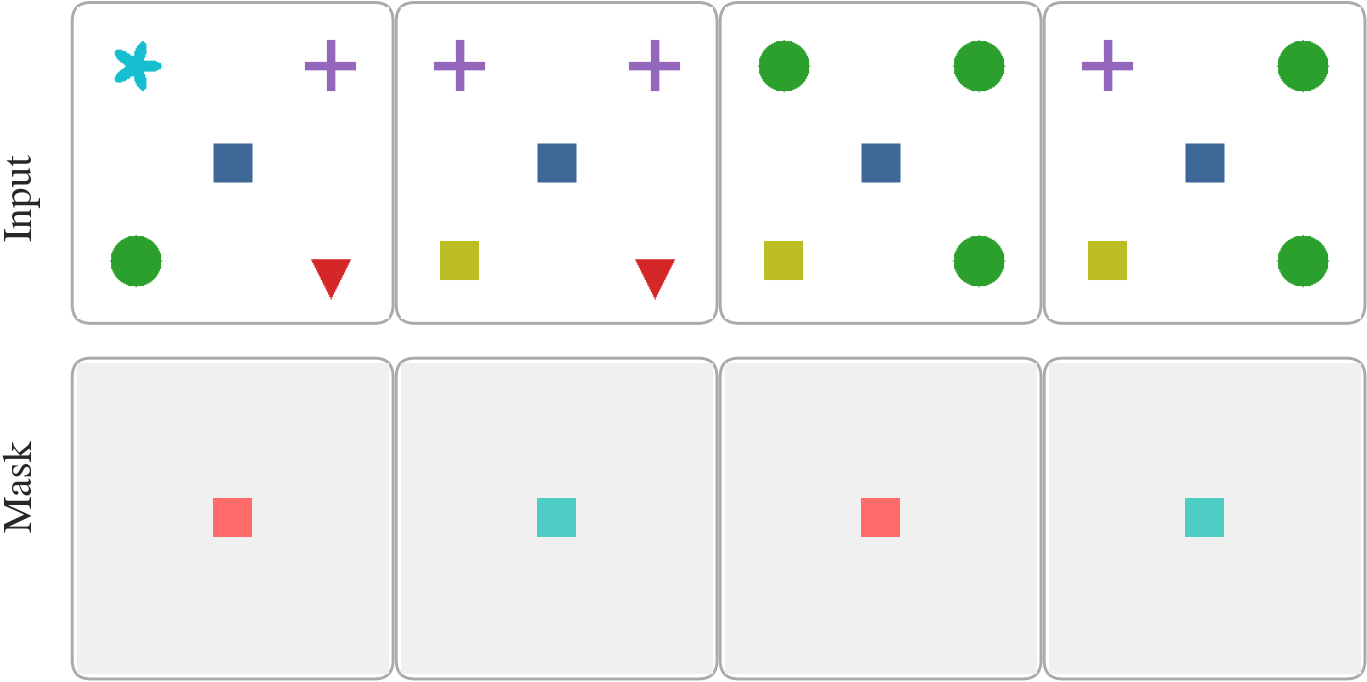}\\[0.4em]
\end{minipage}

\end{table*}
\paragraph{3D Medical Image Segmentation}
To evaluate performance on real-world high-dimensional data requiring long-range spatial understanding, we apply our method to 3D medical image segmentation.
Following the evaluation protocol of \citet{isensee2024nnunetrevisitedrigorousvalidation}, we benchmark on the two datasets identified as most reliable for fair comparisons, namely Kidney and Kidney Tumour Segmentation (\textit{KiTS}), and Automated Cardiac Diagnosis Challenge (\textit{ACDC}). All models are trained and evaluated with identical 5-fold splits, preprocessing/target spacing, augmentations, and postprocessing. Training is conducted under identical conditions, including the same preprocessing and postprocessing steps, allowing observed differences to be attributed solely to the proposed method.
\begin{table}[H]
\centering
\caption{\textbf{3D Medical Image segmentation results} (5-fold CV; mean across 5 folds). Columns report \textbf{DSC} and \textbf{NSD} (at 2\,mm). ``RT'' is runtime (hours) and ``VRAM'' is peak memory (GB). Literature results are shown in gray as reported by \cite{isensee2024nnunetrevisitedrigorousvalidation}.}
\label{tab:main3d}
\small
\setlength{\tabcolsep}{4pt} %
\resizebox{.8\linewidth}{!}{
\begin{tabular}{l*{4}{c}|*{3}{c}}
\toprule
& \multicolumn{2}{c}{\textbf{KiTS}}  & \multicolumn{2}{c}{\textbf{ACDC}} & &&\\ 
\cmidrule(lr){2-3}\cmidrule(lr){4-5}
\textbf{Method} & DSC & NSD & DSC & NSD  & \textbf{Params (M)} & \textbf{RT (h)} & \textbf{VRAM (GB)} \\
\midrule
nnU-Net ResEnc L & 88.98 & 85.74 & 91.40 & 96.21 & 31.12 & 34 & 23 \\
\textbf{SonicNet (Ours)} & 88.55 & 81.19 & 92.02 & 96.07 & 2.59 & 67 & 61.37 \\
\midrule
\rowcolor[gray]{0.96}nnU-Net ResEnc\ L & 88.17 & 85.93   & 91.69 & 95.11 & 31.12 & 36  & 36.60 \\
\rowcolor[gray]{0.96}MedNeXt L k5        & 87.74 & 85.67  & 92.62 & 96.09 & 55.00 & 233 & 18.00 \\
\rowcolor[gray]{0.96}STU-Net L           & 85.84 & 83.02 & 89.34 & 95.12 & 440.30 & 51  & 26.50 \\
\rowcolor[gray]{0.96}SwinUNETRV2         & 84.14 & 80.11 & 86.24 &  95.15 & 72.80 & 15  & 13.40 \\
\rowcolor[gray]{0.96}nnFormer            & 75.85 & 69.43 & 81.55  & 95.83 & 150.0 & 8   & 5.70  \\
\rowcolor[gray]{0.96}CoTr                & 84.59 & 80.92 & 88.02  & 93.74 & 41.27 & 18  & 8.20  \\
\rowcolor[gray]{0.96}U-Mamba Bot        & 86.22 & 83.27 & 89.13 & 95.40 & 64.00 & 24  & 12.40 \\
\bottomrule
\end{tabular}
}
\vspace{-0.75em}
\end{table}
\paragraph{External validation} External validation is critical in medical imaging because models frequently degrade when exposed to new scanners or protocols. Using identical training conditions and evaluating on heterogeneous external datasets provides a clinically meaningful measure of generalisability and highlights whether an architecture is suited for deployment beyond the development cohort. For this generalisability experiment, we evaluate SONIC on the PI-CAI challenge (clinically significant prostate cancer segmentation) data \citep{saha_2022_6624726} and compare it to the top-performing baseline, nnU-Net, on their performance on two external datasets, Prostate158 \citep{ADAMS2022105817} and PROMIS \citep{AHMED2017815}. Qualitive comparison can be found in the appendix.
\begin{center}
    \captionof{table}{\textbf{External validation performance on Prostate158 and PROMIS.}
    SonicNet achieves improved detection performance with substantially fewer parameters.}

\begin{tabular}{
  l
  l
  S[table-format=2.3]
  S[table-format=2.3]
}
\toprule
& \textbf{Metric} & \textbf{nnU-Net} & \textbf{SonicNet} \\
\midrule
\rowcolor{gray!10}
& \multicolumn{1}{l}{\textsc{Trainable Parameters (M/MB)}} &
{31.20/342.0} & {2.59/28.4\textsuperscript{*}} \\
\midrule
\multirow{6}{*}{\rotatebox[origin=c]{90}{\textbf{Prostate158}}}
& AUROC         & 0.814 & \textbf{0.841} \\
& AP            & 0.533 & \textbf{0.548} \\
& F1 Score      & 0.632 & \textbf{0.649} \\
& Sensitivity   & 0.475 & \textbf{0.495} \\
& Precision     & 0.941 & \textbf{0.943} \\
& TP/FP/FN (\%) & {0.30/0.02/0.34} & {0.32/0.02/0.32} \\
\midrule
\multirow{6}{*}{\rotatebox[origin=c]{90}{\textbf{PROMIS}}}
& AUROC         & 0.646 & \textbf{0.687} \\
& AP            & 0.195 & \textbf{0.258} \\
& F1 Score      & 0.185 & \textbf{0.223} \\
& Sensitivity   & 0.103 & \textbf{0.127} \\
& Precision     & \textbf{0.912} & 0.907 \\
& TP/FP/FN (\%) & {0.05/0.01/0.47} & {0.07/0.01/0.47} \\
\bottomrule
\end{tabular}
\end{center}

\paragraph{ImageNet-50M}
To evaluate SONIC on natural images, under highly anisotropic visual conditions, we conduct experiments on ImageNet-1K, the standard benchmark for large-scale image classification. Due to computational constraints, we adopt a reduced training setting that we denote corresponding to 200k optimization steps with a batch size of 256. We evaluate ResNet-50 variants augmented with different spectral operators and compare them against strong baselines, including a Vision Transformer. Beyond reporting standard classification accuracy, we also assess robustness under controlled resolution shifts, which serve as a proxy for the anisotropic distortions common in practical deployment scenarios. By systematically varying input resolution, we quantify how well SONIC maintains accuracy relative to competing methods, thereby characterizing its robustness to scale changes and sampling artifacts.
\begin{figure}[H]
\centering

\begin{minipage}[t]{0.53\textwidth}
    \centering
    \vspace{0pt}%
    \captionof{table}{\textbf{Comparison of ResNet-50 variants} and related architectures on ImageNet under $224\times224$ evaluation.}
    \setlength{\tabcolsep}{6pt} %
   \resizebox{1\linewidth}{!}{%
    \begin{tabular}{@{}lccccc@{}}
        \toprule
        & \multicolumn{3}{c}{\textbf{Model Complexity}} 
        & \multicolumn{2}{c}{\textbf{Accuracy}} \\
        \cmidrule(lr){2-4} \cmidrule(lr){5-6}
        \textbf{Model} & Params & GFLOPs & Img/s & \textbf{Top-1} & \textbf{Top-5} \\
         & (M) &(G) &  &  & \ \\
        \midrule
        ResNet-50         & 25.60 &  8.26 & 4758 & 58.47 & 82.68 \\
        ViT-S/16          & 48.60 & 35.21 & 1136 & 62.23 & 83.91 \\
        \midrule
        ResNet-50 NIFF    & 18.61 & 14.89 &  862 & 57.52  & 82.24   \\
        ResNet-50 S4ND    & 16.67 &  4.57 & 1421 & 64.38 & 86.44 \\
        ResNet-50 GFNet   & 15.72 &  4.57 & 4504 & 61.43 & 84.47 \\
        ResNet-50 RepLK   & 19.23 &  7.71 & 1884 & 65.17  & 86.34   \\
        ResNet-50 Dilated & 25.55 & 38.36 & 2130 & 61.52 & 84.73 \\
        \textbf{ResNet-50 Sonic}  &  1.34     & 0.81 &  831  &  60.01   & 82.28
   \\
        \bottomrule
        \vspace{2mm}
    \end{tabular}
     }
    
    \label{tab:resnet_imagenet_stylish_full}
\end{minipage}
\hfill
\begin{minipage}[t]{0.43\textwidth}
    \centering
    \vspace{0pt}%
    \begin{tikzpicture}
        \begin{axis}[
            width=1\textwidth,
            height=0.77\textwidth,
            xmin=90, xmax=230,
            xtick={96,128,160,192,224},
            xticklabel style={font=\small},
            ymajorgrids=true,
            grid style={dashed,gray!40},
            xlabel={Input resolution (pixels)},
            ylabel={Relative Top-1 degradation (\%)},
            xlabel style={font=\small},
            ylabel style={font=\small},
            ticklabel style={font=\small},
            legend style={
                font=\small,
                at={(0.55,0.45)},
                anchor=north west,
                draw=none,
                fill=none
            },
            line width=1pt,
            mark size=2.2pt,
            y dir=reverse, %
        ]

            \addplot+[mark=o]
                coordinates {
                    (96,  30.38)
                    (128, 13.35)
                    (160, 5.97)
                    (192, 2.50)
                    (224, 0.00)
                };
            \addlegendentry{ResNet-50}

            \addplot+[mark=square*]
                coordinates {
                    (96,  24.26)
                    (128, 13.60)
                    (160, 5.38)
                    (192, 4.74)
                    (224, 0.00)
                };
            \addlegendentry{NIFF}

            \addplot+[mark=halfcircle*]
                coordinates {
                    (96,  21.52)
                    (128, 14.50)
                    (160, 3.23)
                    (192, 2.94)
                    (224, 0.00)
                };
            \addlegendentry{Global Filter}

            \addplot+[mark=triangle*]
                coordinates {
                    (96,  20.09)
                    (128, 8.87)
                    (160, 2.23)
                    (192, 1.73)
                    (224, 0.00)
                };
            \addlegendentry{SONIC}

        \end{axis}
    \end{tikzpicture}
    \captionof{figure}{Relative performance degradation under resolution changes on ImageNet.}
    \label{fig:resolution_degradation}
\end{minipage}

\vspace{-5mm}
\end{figure}

\paragraph{Compute and memory overhead.}
Figure 4 illustrates the compute and memory profile of SONIC in comparison to a standard $3{\times}3$ convolution and a ViT block with $4{\times}4$ patches. At scale ($224 \times 224$), SONIC is only $1.23{\times}$ slower and uses $1.18{\times}$ more memory, representing a modest overhead for obtaining global receptive fields. At higher resolutions, the runtime gap narrows and SONIC becomes effectively on par with
convolution, reflecting the favourable scaling
of FFT-based filtering. SONIC's peak memory is dominated by the FFT stage, which requires storing the full spectral grid at each layer. In contrast, full-resolution self-attention grows quadratically with spatial size,
becoming substantially more expensive even at moderate resolutions. Furthermore, as shown in Fig \ref{fig:influence_c_m} (appendix), runtime and memory grow approximately linearly in both $C$ and $M$ in 
the practically relevant regime, with no unexpected spikes. This confirms that SONIC can be tuned in the number of channels and number of modes without additional overheads. Overall, SONIC provides global spatial mixing at a fraction of the cost of global attention, while remaining close to convolution in both compute and memory. 
\begin{figure}[H]
\centering

\begin{minipage}[t]{0.40\textwidth}
\centering
\vspace{0pt}

\begin{tikzpicture}
\begin{axis}[
    width=1.0\textwidth,
    height=0.85\textwidth,
    xlabel={Resolution $H = W$ (pixels)},
    ylabel={Runtime per block (ms)},
    xmajorgrids=true,
    ymajorgrids=true,
    grid style={dashed,gray!40},
    ymode=log,
    xmin=60, xmax=520,
    xtick={64,96,128,160,192,224,256,320,384,448,512},
    xticklabels={64,96,128,160,192,224,256,320,384,448,512},
    ticklabel style={font=\footnotesize},
    xticklabel style={font=\footnotesize, rotate=45, anchor=east}, %
    xlabel style={font=\small},
    ylabel style={font=\small},
    legend style={
        font=\small,
        at={(0.0,0.99)},
        anchor=north west,
        draw=none
    },
    mark size=2.2pt,
    line width=1pt
]

\addplot+[mark=triangle*, color=blue]
coordinates {
    (64,8.2573) (80,9.1512) (96,10.1062) (112,12.289)
    (128,14.7959) (144,19.525) (160,24.780)
    (176,33.3095) (192,42.090) (224,69.8197)
    (256,111.289) (288,172.2398) (320,256.009)
    (384,523.327) (448,949.745) (512,1593.746)
};
\addlegendentry{ViT block};

\addplot+[mark=o, color=orange]
coordinates {
    (64,4.0190) (80,5.9147) (96,8.2021) (112,10.869)
    (128,13.705) (144,17.354) (160,21.017)
    (176,25.182) (192,29.765) (224,40.2317)
    (256,52.135) (288,65.787) (320,81.043)
    (384,116.133) (448,157.549) (512,205.822)
};
\addlegendentry{Conv block};

\addplot+[mark=square*, color=green!60!black]
coordinates {
    (64,17.4565) (80,18.8546) (96,20.9506) (112,23.0006)
    (128,25.8288) (144,28.988) (160,31.988)
    (176,35.3989) (192,39.514) (224,49.50415)
    (256,58.957) (288,72.164) (320,86.547)
    (384,119.298) (448,158.199) (512,198.792)
};
\addlegendentry{SONIC block};

\end{axis}
\end{tikzpicture}

\caption*{(a) Runtime per block (log scale)}
\end{minipage}
\hspace{10pt}
\begin{minipage}[t]{0.40\textwidth}
\centering
\vspace{0pt}

\begin{tikzpicture}
\begin{axis}[
    width=1.0\textwidth,
    height=0.85\textwidth,
    xlabel={Resolution $H = W$ (pixels)},
    ylabel={Runtime per block (ms)},
    xmajorgrids=true,
    ymajorgrids=true,
    grid style={dashed,gray!40},
    ymode=log,
    xmin=60, xmax=520,
    xtick={64,96,128,160,192,224,256,320,384,448,512},
    xticklabels={64,96,128,160,192,224,256,320,384,448,512},
    ticklabel style={font=\small},
    xticklabel style={font=\small, rotate=45, anchor=east}, %
    xlabel style={font=\small},
    ylabel style={font=\small},
    legend style={
        font=\small,
        at={(0.44,0.35)},
        anchor=north west,
        draw=none
    },
    mark size=2.2pt,
    line width=1pt
]

\addplot+[mark=triangle*, color=blue]
coordinates {
    (64,340.7) (80,496.9) (96,687.3) (112,910.9)
    (128,1170.2) (144,1464.0) (160,1792.3)
    (176,2155.2) (192,2552.6) (224,3451.3)
    (256,4488.1) (288,5663.3) (320,6976.7)
    (384,10018.9) (448,13613.3) (512,17760.2)
};
\addlegendentry{ViT block};

\addplot+[mark=o, color=orange]
coordinates {
    (64,578.5) (80,866.5) (96,1218.5) (112,1634.5)
    (128,2114.5) (144,2658.5) (160,3266.5)
    (176,3938.5) (192,4674.5) (224,6338.5)
    (256,8258.5) (288,10434.5) (320,12866.5)
    (384,18498.5) (448,25154.5) (512,32834.5)
};
\addlegendentry{Conv block};

\addplot+[mark=square*, color=green!60!black]
coordinates {
    (64,677.9) (80,1020.8) (96,1435.7) (112,1927.4)
    (128,2494.1) (144,3136.0) (160,3853.0)
    (176,4645.4) (192,5512.5) (224,7473.0)
    (256,9733.3) (288,12295.5) (320,15156.7)
    (384,21781.5) (448,29609.0) (512,38638.8)
};
\addlegendentry{SONIC block};

\end{axis}
\end{tikzpicture}

\caption*{(b) Peak GPU memory (log scale)}
\end{minipage}

\caption{Runtime and memory characteristics of ViT, convolutional, and SONIC blocks across spatial resolutions.}
\end{figure}

\section{Discussion}
We introduced a spectral factorisation framework, where SONIC serves as a theoretically grounded alternative to spatial convolution blocks. Unlike conventional spatial kernels, SONIC employs low-rank, orientation-aware operators in the frequency domain. This design provides a principled inductive bias for modelling long-range, structured interactions while remaining highly parameter-efficient. 
Our empirical evaluation demonstrates SONIC's properties. On SynthShape, the model exhibited superior robustness to image distortions compared to conventional CNN and ViT baselines and previous spectral-domain architectures. In the HalliGalli spatial reasoning task, SONIC was the only architecture capable of solving strict long-range dependencies within a single block, highlighting the effectiveness of its global receptive field. Furthermore, these theoretical advantages translated into real-world performance in 3D medical segmentation benchmarks (KiTS and ACDC), where SONIC matched or exceeded state-of-the-art performance while requiring significantly fewer parameters ($<10\%$) than established heavyweights such as nnU-Net and MedNeXt.

At the same time, important limitations remain. Nonlinearities must be applied in the spatial domain. This prevents us from stacking multiple SONIC layers purely in the frequency domain and forces repeated FFT/IFFT operations, which introduce additional overhead. Although this limitation is shared by most spectral neural architectures, it does constrain how fully the model can operate within the spectral domain. Furthermore, we observed occasional instabilities during SONIC block initialisation, stemming from the same property that defines the operator: in imaging tasks, identical spatial dimensions may correspond to very different physical scales across datasets. Developing a more general and robust initialisation scheme for SONIC, therefore, remains an important direction for future work. Moreover, the global nature of the frequency-domain representation can limit the capture of very fine local structure, which suggests that hybrid architectures may ultimately be needed to combine the strengths of both domains. 
Our goal here is to provide SONIC as a general and simple operator that can be integrated in the same way as other well-known alternatives. Further work should explore how to incorporate SONIC thoughtfully into new or existing architectures. In summary, spectral factorisation offers a new building block for neural architectures that complements existing paradigms. Its strengths lie in long-range receptive field, parameter efficiency, orientation-awareness, and robustness, while future work should focus on improving efficiency, mitigating memory demands, and exploring hybrid spectral-spatial architectures.

\section{Acknowledgements}
The authors would like to acknowledge the Research High Performance Computing (RHPC) facility of the Netherlands Cancer Institute (NKI).
In this paper, we used large language models to refine wording and improve the clarity of information transfer. All conceptual ideas, discussion of related work, and factual content were developed manually by the authors; the models were employed solely for assistance with presentation.

\bibliography{iclr2026_conference}
\bibliographystyle{iclr2026_conference}
\newpage
\section{appendix}
\section*{Appendix A: Implementation details}
We constrain $s_m>0$ and typically enforce $\operatorname{Re(a_m)<0)}$ so that the spatial response
function decays rather than grows. The imaginary part $\operatorname{Im(a_m)}$ can be bounded in
magnitude (e.g., $|a^{\mathrm{im}}_m|\le \rho$). We initialize $v_m \sim U(0, \pi)$.\\ \\
All parameters are learned end-to-end by backpropagation. A convenient
reparameterisation that enforces the constraints is:
\[
s_m = \mathrm{softplus}(\sigma_m) + \varepsilon, \qquad
a^{\mathrm{re}}_m = -\,\mathrm{softplus}(\alpha_m),
\qquad
a^{\mathrm{im}}_m = \rho \,\tanh(\beta_m),
\qquad
v_m = \frac{u_m}{\|u_m\|_2},
\]
with free variables $\sigma_m,\alpha_m,\beta_m,\rho \in \mathbb{R}$ and $u_m\in\mathbb{R}^2$,
small $\varepsilon>0$.
The mixing matrices $B$ and $C$ are complex-valued and learned without constraints.

\noindent\textit{Implementation notes.}
(i) We standardize each input channel to zero mean and unit variance, with a small noise for numerical stability.
(ii) We apply an RMS transfer gain normalisation over the (half-)spectrum to keep the overall response well-scaled across resolutions.(iii) We use real–FFT ($\mathrm{rFFT}$/$\mathrm{irFFT}$) along the last two spatial dimensions; consequently we enforce Hermitian consistency by zeroing the imaginary part at DC.(iii) For memory efficiency the computation is performed in frequency \emph{slabs} (blocks over rows of $\Omega$) without altering the continuous formulation above.  
(iv) Direction vectors are rescaled by $D_\Delta^{-1}$ and renormalized (unit length) before use, ensuring invariance to pixel spacing.  (v) Optional mode dropout is applied to $V_m$ as a regularizer.
\subsubsection*{SynthShape}
\label{appendix:implementation_gs}
The dataset consist of a random number of geometric primitives (circle, square, triangle, cross, star) at random positions and scales within the image, while preventing overlaps through collision checks. Each object is assigned a randomly perturbed base colour, ensuring that models cannot exploit a trivial mapping between RGB values and semantic classes. The ground-truth segmentation mask assigns a unique class label to each shape type, with background indexed as class 0.  

\paragraph{Models.}
All models use an embedding width of $c{=}128$ 
\begin{itemize}
    \item \textbf{ConvNet:} A lightweight stack of $L$ convolutional layers (default $L=4$), each followed by group normalisation and GELU activations. A $1\times1$ convolution projects the final feature map to the number of classes. The patch size is set to 16 to give the model a fair opportunity to capture broader context, rather than learning solely from small local receptive fields.
    \item \textbf{ViT:} A Vision Transformer consisting of a patch embedding layer, sinusoidal positional encodings (interpolated if image resolution differs), and a stack of transformer blocks with multi-head self-attention and MLP layers. The output features are reshaped and upsampled to the original spatial resolution, followed by a $1\times1$ convolution for classification.  
    \item \textbf{SonicNet:} For SonicNet we use a depth of $4$ stacked SonicBlocks, each consisting of GroupNorm, GELU, and a residual spectral convolutional mapping. The final stage applies GroupNorm, GELU, and a $3{\times}3$ convolution to project features to class logits. 
     \item \textbf{GFNet:} Each block replaces local convolutions by a learned complex-valued mask applied in the Fourier domain. Features are normalised and transformed by the global filter, followed by a pointwise MLP and residual connections, while the overall encoder–head structure is kept identical to the ConvNet.
    \item \textbf{NIFF:} Rach block learns a continuous frequency response via a small MLP that maps frequency coordinates to complex filter values. These filters are applied depthwise in the Fourier domain and wrapped in the same normalisation, residual, and head structure as the ConvNet.
 \item \textbf{S4ND:} A state-space baseline where the convolutional backbone is replaced by stacked S4ND layers operating directly on the $H\times W$ grid. Each block applies a 2D structured state-space update to the feature map and is embedded in the same residual and segmentation head pattern as the other models.

\end{itemize}

\paragraph{Training.}  
All models were trained using the AdamW optimizer with learning rate $10^{-2}$ and weight decay $10^{-4}$, for $1000$ epochs and batch size $32$. A one-cycle learning rate schedule was applied. To account for class imbalance, inverse-frequency class weights were computed dynamically from a large synthetic batch and used in the cross-entropy loss. The final training objective combined cross-entropy with the multi-class Dice loss in equal weighting. 

\paragraph{Evaluation.}  
Model robustness was assessed by applying five geometric transformations at inference: rescaling, rotation, translation, distortion, and Gaussian noise. Each transformation was applied with three levels of severity. Rescaling resized the full image before resampling it back to $64\times64$, introducing interpolation artefacts. Translation shifted the input by a fixed percentage of image width/height, potentially moving parts of objects out of frame. Distortion was implemented via bicubic upsampling of a low-resolution displacement field. Rotation was performed around the image centre, and Gaussian noise was added per pixel channel.  

\paragraph{Metrics.}  
The primary evaluation metric was the multi-class Dice score (excluding background), averaged across folds. All experiments were repeated 5 times with different seeds to estimate variance.  
\ref{tab:geosynth_full_results}.
\begin{figure}[H]
        \centering
        \includegraphics[width=.5\linewidth]{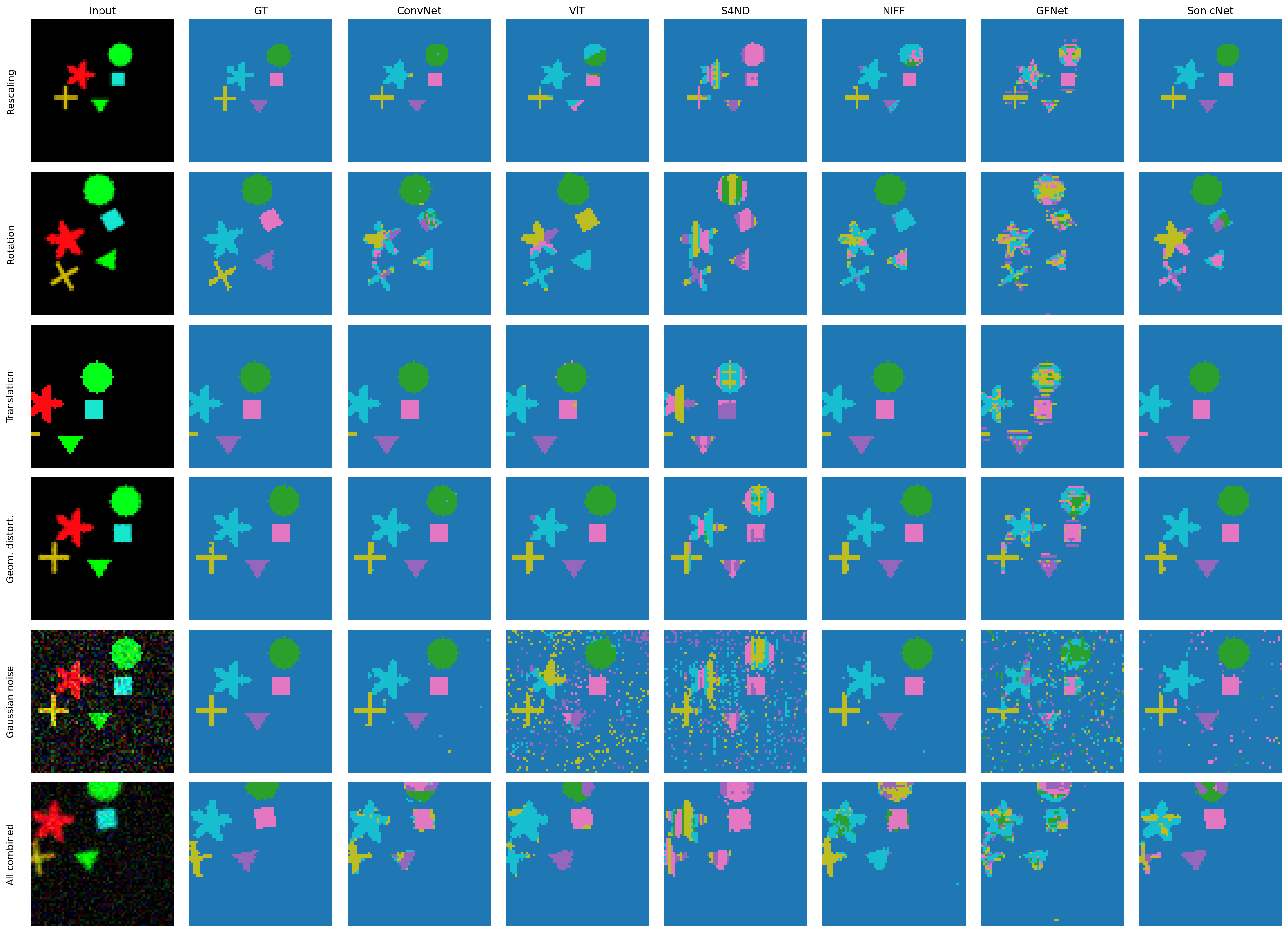}
        \caption{Qualitative visual examples on the SynthShape benchmark}
        \label{fig:conv_result}
\end{figure}\vspace{-5mm}
\subsubsection*{ImageNet}
For ImageNet we follow the public ``academic default'' implementation\footnote{\url{https://github.com/landskape-ai/imagenet}} and keep all training hyperparameters and optimisation settings unchanged. We replace the standard ResNet-50 bottleneck blocks by SonicBlocks, where each $3{\times}3$ convolution in the main path is substituted with a Sonic layer, while the $1{\times}1$ convolutions in the skip path and classifier head remain unchanged. Full architectural details and the exact PyTorch implementation of \texttt{resnet50\_sonic} are provided in the supplementary material.

\subsubsection*{Medical Imaging Benchmark}
\paragraph{Setup.}
Following the recommendations of \citet{isensee2024nnunetrevisitedrigorousvalidation}, we minimize confounding factors and keep the experiment as plain as possible. We retain the baseline nnU-Net preprocessing and postprocessing and change only the network backbone: the original U-Net is replaced by a stack of SONIC Blocks (``SonicNet''). The first block lifts the input from $C$ to $K$ channels; the remaining $D-1$ blocks keep $K$ channels. We apply GroupNorm and GELU before a final $3{\times}3$ convolution to produce $n_{\text{classes}}$ output channels. For this experiment, we used four stacked SonicBlocks (i.e., a depth of 4). 

We employed stochastic gradient descent with an initial learning rate of $10^{-2}$ and a weight decay of $10^{-5}$. Training was performed with a mini-batch size of two for a total of 1000 epochs, each consisting of 250 iterations. For inference, we used the checkpoint corresponding to the highest validation performance during training.
\subsection{Qualitative comparison of the external validation}
\begin{figure}[H]
\centering
\begin{minipage}{\textwidth}
    \centering

    \includegraphics[width=0.24\linewidth]{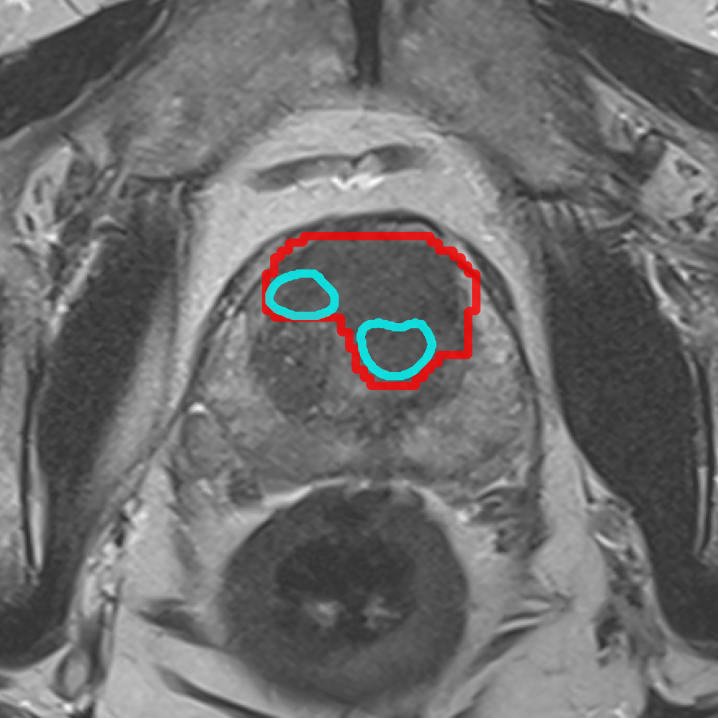}\hfill
    \includegraphics[width=0.24\linewidth]{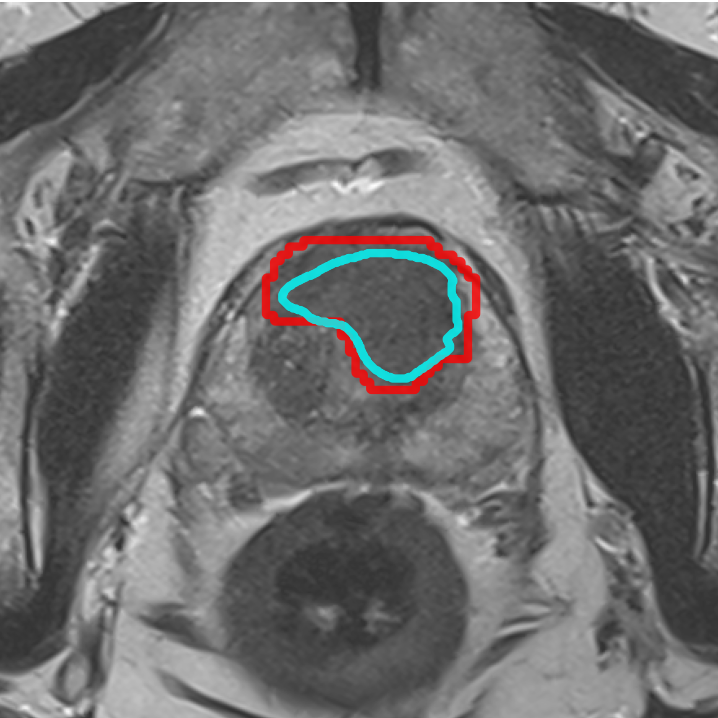}\hfill
    \includegraphics[width=0.24\linewidth]{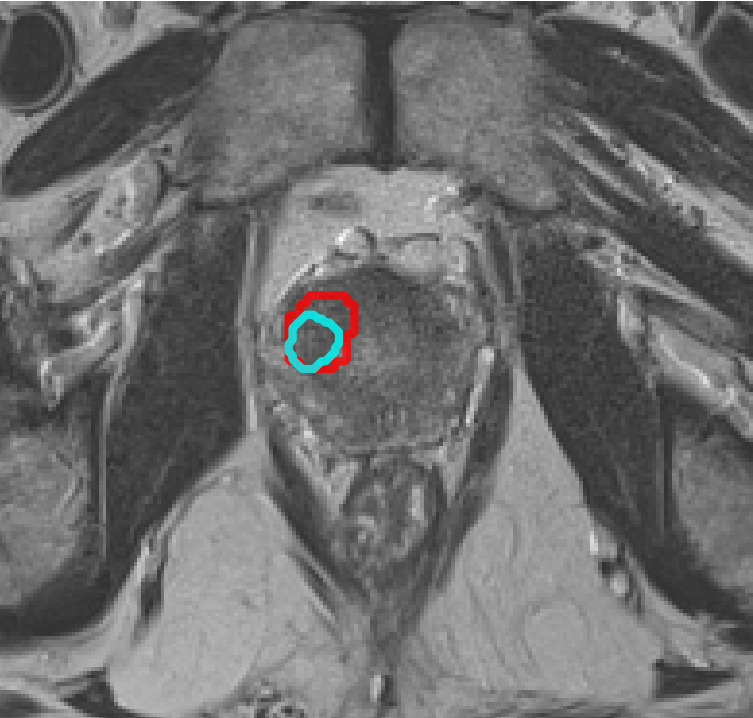}\hfill
    \includegraphics[width=0.24\linewidth]{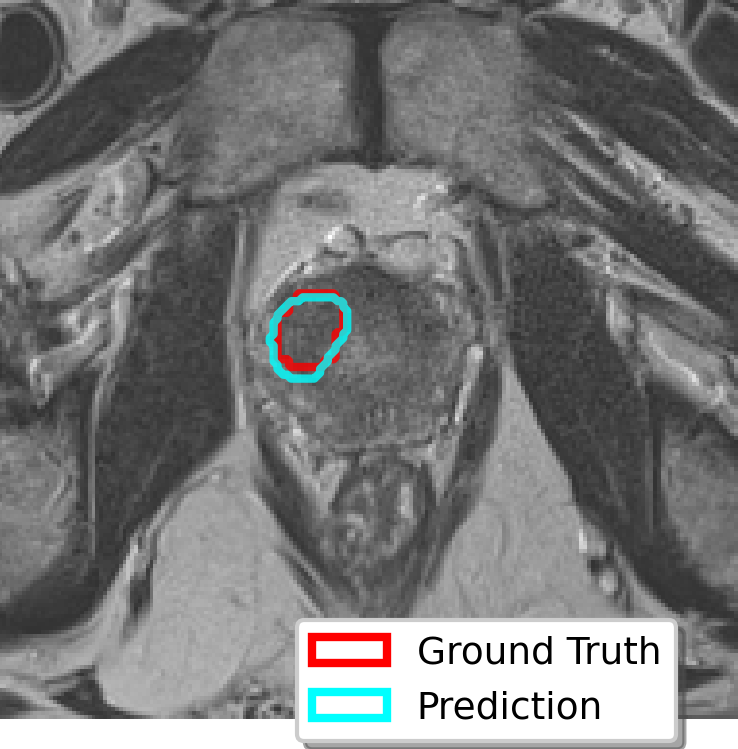}

    \captionof{figure}{\textbf{Qualitative comparison of prostate cancer detection methods.}
    The figure shows representative cases from the Prostate158 (left) and PROMIS (right)
    datasets, with ground truth lesions (red) and model predictions (cyan) overlaid on
    T2-weighted MRI slices (confidence $\geq 0.5$).}
    \label{fig:prostate_qualitative}
\end{minipage}
\end{figure}

\section{Appendix B: Practical implementation}
\paragraph{Role of \texorpdfstring{$K$}{K} and \texorpdfstring{$M$}{M}}
The parameters $K$ and $M$ play complementary roles in shaping the behaviour of a SONIC block. The number of modes $M$ determines the spectral diversity of the operator; in contrast, the channel width $K$ controls the capacity with which these shared modes are mixed across feature channels. The ratio between $K$ and $M$ therefore reflects the balance between channel-mixing capacity and spectral richness. Understanding this trade-off helps guide architectural choices across different model sizes.
\paragraph{Qualitative analysis of receptive fields} 
To better understand how SONIC behaves in practice, we visualize the normalized spectral energy of the learned filters across the four stages of the network. Each plot shows the log-scaled energy distribution over the spatial frequency plane, giving an intuitive sense of the effective receptive field and directional structure captured at different depths.

As the network progresses through stages, the spectral responses become increasingly smooth, structured, and oriented—indicating that early stages capture broad, irregular frequency content, while deeper stages refine this into cleaner, more coherent spectral patterns. These visualizations highlight how SONIC gradually organizes its spectral modes and how the parameterization remains stable and well-behaved across depth.
\begin{figure}[H]
    \centering
    \includegraphics[width=0.5\linewidth]{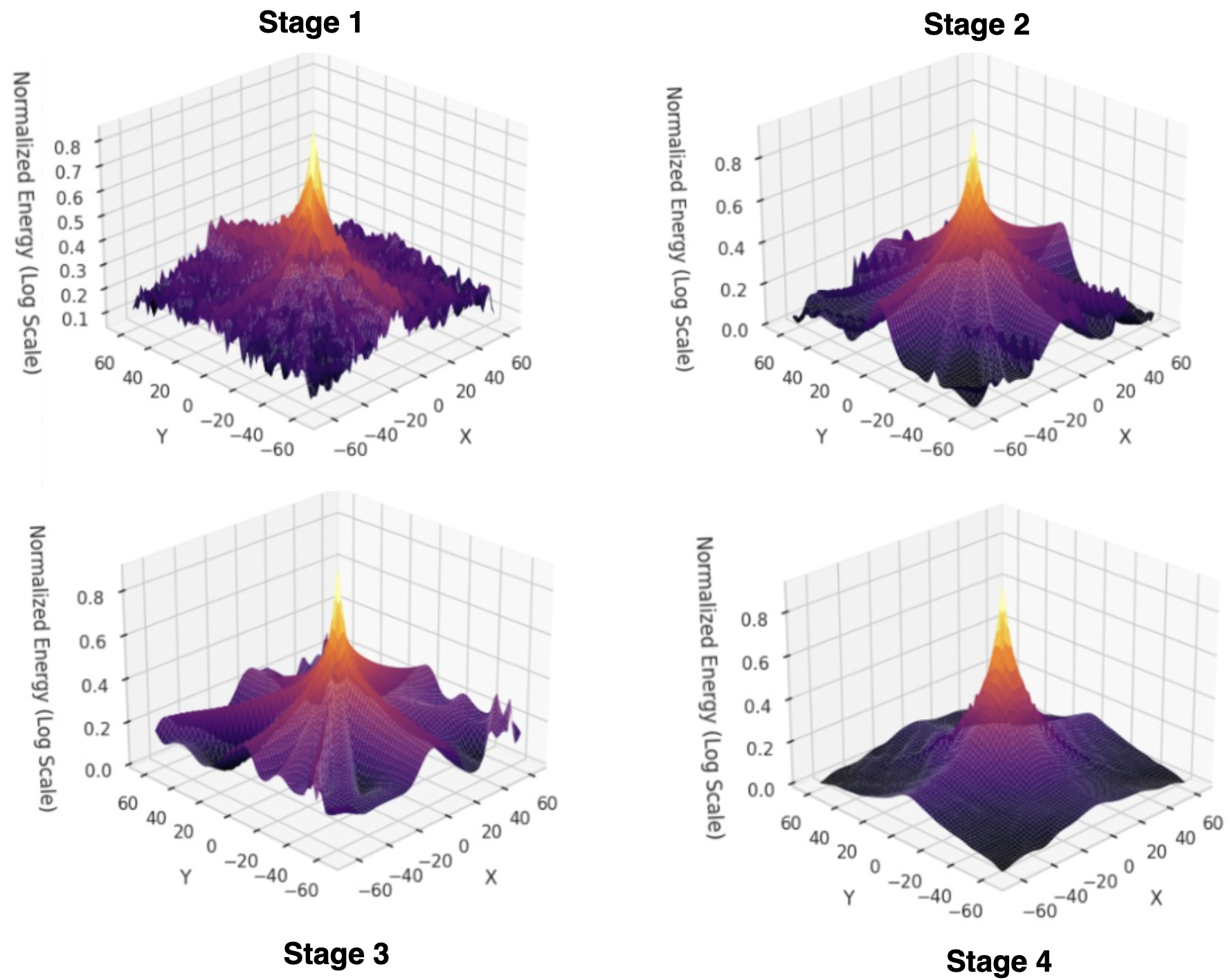}
    \caption{Visualisation of a randomly selected learned 2D convolutional kernel from our medical image segmentation model across four stages of the network.}
    \label{fig:overhead}
\end{figure}
\paragraph{Practical scalability of SONIC Block}
To validate that the SONIC block exhibits the intended linear scaling behavior, we empirically benchmark its runtime and memory usage across a range of channel dimensions and mode counts.We confirm this behavior by measuring wall-clock runtime and peak memory under controlled synthetic settings, sweeping C and M independently while keeping spatial resolution fixed. Across all tested configurations, runtime increases as a straight line with respect to both variables, and memory usage follows the same linear trend.
\begin{figure}[H]
    \centering
    \includegraphics[width=0.5\linewidth]{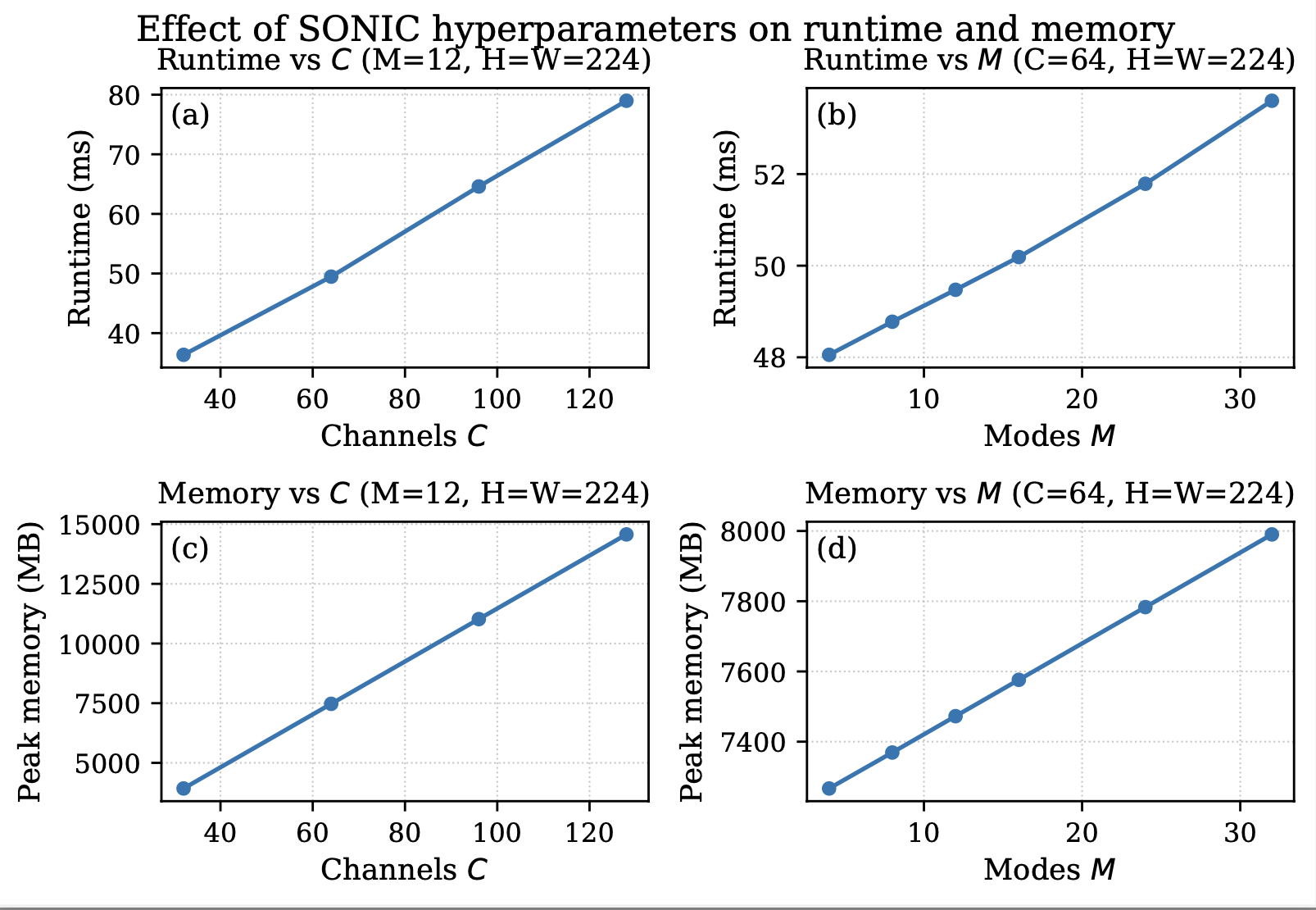}
    \caption{Runtime and memory of SONIC when varying channels $C$ and modes $M$ at fixed resolution.
}
    \label{fig:influence_c_m}
\end{figure}
\section*{Appendix C: Supporting Proofs}
\label{appendix:convolution_theorem}
\subsection*{Convolution Theorem for the $D$-dimensional Fourier Transform}

Let the convolution of two functions on $\mathbb{R}^D$ be defined by
\[
(f*g)(\mathbf{x})
:= \int_{\mathbb{R}^D}
f(\boldsymbol{\tau})\,g(\mathbf{x}-\boldsymbol{\tau})\,d\boldsymbol{\tau},
\qquad \mathbf{x}\in\mathbb{R}^D.
\]
Then, for $\boldsymbol{\omega}\in\mathbb{R}^D$, the $D$-dimensional Fourier transform satisfies
\begin{align*}
\mathcal{F}\{f*g\}(\boldsymbol{\omega})
&= \int_{\mathbb{R}^D}
\Bigg(\int_{\mathbb{R}^D}
f(\boldsymbol{\tau})\,g(\mathbf{x}-\boldsymbol{\tau})\,d\boldsymbol{\tau}\Bigg)
e^{-i\boldsymbol{\omega}\cdot \mathbf{x}}\,d\mathbf{x} \\
&= \int_{\mathbb{R}^D}\int_{\mathbb{R}^D}
f(\boldsymbol{\tau})\,g(\mathbf{u})\,
e^{-i\boldsymbol{\omega}\cdot(\boldsymbol{\tau}+\mathbf{u})}\,d\mathbf{u}\,d\boldsymbol{\tau} \\
&= \Bigg(\int_{\mathbb{R}^D} f(\boldsymbol{\tau})\,e^{-i\boldsymbol{\omega}\cdot \boldsymbol{\tau}}\,d\boldsymbol{\tau}\Bigg)
   \Bigg(\int_{\mathbb{R}^D} g(\mathbf{u})\,e^{-i\boldsymbol{\omega}\cdot \mathbf{u}}\,d\mathbf{u}\Bigg).
\end{align*}

Hence,
\[
\mathcal{F}\{f*g\}(\boldsymbol{\omega})
= \mathcal{F}\{f\}(\boldsymbol{\omega}) \,\mathcal{F}\{g\}(\boldsymbol{\omega}).
\]
\subsection*{Connection to State-Space Kernels}
\label{appendix:LTI_sonic}
Consider the linear time-invariant state-space model
\begin{equation}
\dot x(t)=Ax(t)+Bu(t), \qquad y(t)=Cx(t),
\label{eq:ssm}
\end{equation}
with $x(t)\in\mathbb{C}^n$, $u(t)\in\mathbb{C}^m$, $y(t)\in\mathbb{C}^p$, and system matrices
$A\in\mathbb{C}^{n\times n}$, $B\in\mathbb{C}^{n\times m}$, $C\in\mathbb{C}^{p\times n}$.
Assume a zero initial state $x(0^-)=0$ and a strictly proper output.

The corresponding impulse response (or kernel) is
\begin{equation}
\mathcal{K}(t) = C e^{At} B, \qquad t \geq 0.
\label{eq:continuous_kernel}
\end{equation}
By definition, the transfer function is the Laplace transform of the impulse response:
\begin{equation}
H(s) = \int_0^\infty e^{-st} \, \mathcal{K}(t) \, dt
= \int_0^\infty e^{-st} \, C e^{At} B \, dt.
\end{equation}
Pulling out $C$ and $B$ gives
\begin{equation}
H(s) = C \left( \int_0^\infty e^{(A - sI)t} \, dt \right) B.
\end{equation}
For $\operatorname{Re}(s)$ sufficiently large, the integral converges to
\begin{equation}
\int_0^\infty e^{(A - sI)t} \, dt = (sI - A)^{-1}.
\end{equation}
Hence the transfer function is
\begin{equation}
H(s) = C (sI - A)^{-1} B.
\label{eq:transfer_deriv}
\end{equation}

\paragraph{SONIC with Restricted Modes.}
We show that our general Fourier domain formulation reduces to the Laplace resolvent parameterisation of \textsc{S4ND} when orientations are
restricted to the coordinate axes.

\medskip
Recall our frequency response factorisation
\begin{equation}
\widehat{H}_{c,k}(\boldsymbol{\omega})
= \sum_{m=1}^{M} C_{k m}\, T_m(\boldsymbol{\omega})\, B_{m c},
\end{equation}
with mode response
\begin{equation}
T_m(\boldsymbol{\omega})
= \frac{1}{\, i s_m (\boldsymbol{\omega}\!\cdot\! v_m) \;-\; a_m \;+\; 
\tau_m \|(I - v_m v_m^\top)\boldsymbol{\omega}\|^2 }.
\end{equation}
Suppose $v_m = e_d$, the $d$-th standard basis vector. Then
\[
\boldsymbol{\omega}\!\cdot\! v_m = \omega_d,
\qquad
(I - v_m v_m^\top)\boldsymbol{\omega}
= \sum_{j\neq d} \omega_j e_j,
\]
so that
\[
\|(I - v_m v_m^\top)\boldsymbol{\omega}\|^2
= \sum_{j\neq d} \omega_j^2.
\]

In this case,
\begin{equation}
T_m(\boldsymbol{\omega})
= \frac{1}{\, i s_m \,\omega_d - a_m + \tau_m \sum_{j\neq d} \omega_j^2 }.
\end{equation}

We discard the transverse penalty $\tau_m = 0$, then
\[
T_m(\omega_d)
= \frac{1}{\,i s_m \omega_d - a_m\,}
= \frac{1}{s_m}\,\frac{1}{\,i\omega_d - \frac{a_m}{s_m}\,},
\]
where the absorption is into the learned parameters ($ a_m/s_m$ in $A$, and $B$ or $C$ absorb $1/s_m$).
Thus
\[
\widehat H_{c,k}(\omega_d)
= \big[C\,(i\omega_d I - A)^{-1}B\big]_{kc},
\qquad
H(s)=C(sI-A)^{-1}B.
\]

\end{document}